\documentclass[10pt,twocolumn,letterpaper]{article}

\usepackage{iccv}
\usepackage{times}
\usepackage{epsfig}
\usepackage{graphicx}
\usepackage{amsmath}
\usepackage{amssymb}
\usepackage{booktabs}
\usepackage{multirow}
\usepackage{tabularx}
\usepackage{makecell}
\usepackage{paralist}
\usepackage{subcaption}

\usepackage[pagebackref=true,breaklinks=true,letterpaper=true,colorlinks,bookmarks=false]{hyperref}

\iccvfinalcopy 

\def\iccvPaperID{} 

\newcommand{\valseen}{Val-Seen\xspace}  
\newcommand{\valunseen}{Val-Unseen\xspace}
\newcommand{\pathdreamer}{Pathdreamer\xspace}
\newcommand{\segmodel}{Structure Generator\xspace}  
\newcommand{\rgbmodel}{Image Generator\xspace}  
\newcommand{\threesixty}{$360^{\circ}$\xspace}
\newcommand{\norm}[1]{\left\lVert#1\right\rVert} 

\newcommand{\cutabstractup}{\vspace{-0.1in}}
\newcommand{\cutabstractdown}{\vspace{-0.1in}}

\newcommand{\cutsectionup}{\vspace{-0.05in}}
\newcommand{\cutsectiondown}{\vspace{-0.05in}}

\newcommand{\cutsubsectionup}{\vspace{-0.05in}}
\newcommand{\cutsubsectiondown}{\vspace{-0.05in}}

\newcommand{\cutparagraphup}{\vspace{-0.1in}}

\def\And{
  \end{tabular}\\[1.5ex]%
  \begin{tabular}[t]{c}}

\begin{document}

\title{Pathdreamer: A World Model for Indoor Navigation}


\author{%
Jing Yu Koh$^1$
\and
Honglak Lee$^2$
\and
Yinfei Yang$^1$
\and
Jason Baldridge$^1$
\and
Peter Anderson$^1$
\And
$^1$Google Research\and
$^2$University of Michigan
}

\maketitle

\begin{abstract}
\cutabstractup
People navigating in unfamiliar buildings take advantage of myriad visual, spatial and semantic cues to efficiently achieve their navigation goals. Towards equipping computational agents with similar capabilities, we introduce \pathdreamer, a visual world model for agents navigating in novel indoor environments. Given one or more previous visual observations, \pathdreamer  generates plausible high-resolution \threesixty visual observations (RGB, semantic segmentation and depth) for viewpoints that have not been visited, in buildings not seen during training. In regions of high uncertainty (e.g. predicting around corners, imagining the contents of an unseen room), \pathdreamer can predict diverse scenes, allowing an agent to sample multiple realistic outcomes for a given trajectory. We demonstrate that \pathdreamer encodes useful and accessible visual, spatial and semantic knowledge about human environments by using it in the downstream task of Vision-and-Language Navigation (VLN). Specifically, we show that planning ahead with \pathdreamer brings about half the benefit of looking ahead at actual observations from unobserved parts of the environment. We hope that \pathdreamer will help unlock model-based approaches to challenging embodied navigation tasks such as navigating to specified objects and VLN.

\cutabstractdown
\end{abstract}

\cutsectionup
\section{Introduction}
\cutsectiondown

\begin{figure}[t]
\begin{center}
\includegraphics[width=1.0\linewidth]{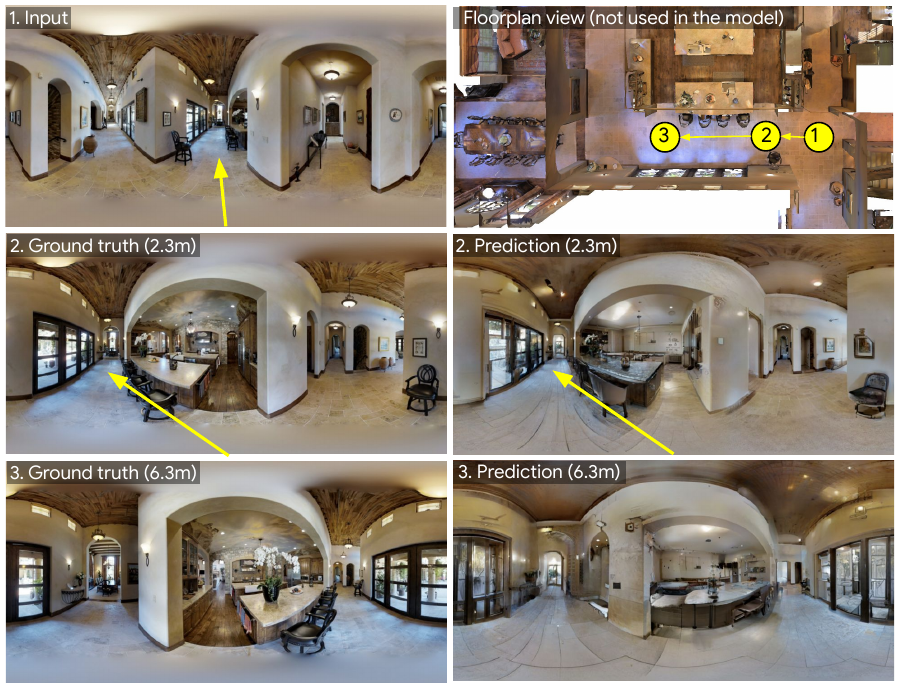}
\end{center}
\vspace{-0.2in}
\caption{Generating photorealistic \threesixty visual observations from an imagined 6.3m trajectory in a previously unseen building. Observations also include depth and segmentations (not shown here).}
\vspace{-0.2in}
\label{fig:teaser}
\end{figure}

World models \cite{ha2018world}, or models of environments \cite{suttonbarto2018}, are an appealing way to represent an agent's knowledge about its surroundings. An agent with a world model can predict its future by `imagining' the consequences of a series of proposed actions. This capability can be used for sampling-based planning~\cite{finn2017deep, nagabandi2018neural}, learning policies directly from the model (i.e., learning in a dream)~\cite{finn2016deep,ha2018world,piergiovanni2019learning,hafner2020mastering}, and for counterfactual reasoning~\cite{buesing2018woulda}. Model-based approaches such as these also typically improve the sample efficiency of deep reinforcement learning~\cite{suttonbarto2018,osinski20}. However, world models that generate high-dimensional visual observations (i.e., images) have  typically been restricted to relatively simple environments, such as Atari games~\cite{osinski20} and tabletops~\cite{finn2017deep}.

Our goal is to develop a generic visual world model for agents navigating in indoor environments. Specifically, given one or more previous observations and a proposed navigation action sequence, we aim to generate plausible high-resolution visual observations for viewpoints that have not been visited, and do so in buildings not seen during training.
Beyond applications in video editing and content creation, solving this problem would unlock model-based methods for many embodied AI tasks, including navigating to objects~\cite{batra2020objectnav}, instruction-guided navigation~\cite{anderson2018vision, Qi_2020_CVPR, ku2020room} and dialog-guided navigation~\cite{thomason2020vision,hahn2020way}. 
For example, an agent asked to find a certain type of object in a novel building, e.g. `\textit{find a chair}', could perform mental simulations using the world model to identify navigation trajectories that are most likely to include chair observations -- without moving.

Building such a model is challenging. It requires synthesizing completions of partially visible objects, using as few as one previous observation. This is akin to novel view synthesis from a single image~\cite{flynn2016deep,wiles2020synsin}, but with potentially unbounded viewpoint changes. There is also the related but considerably more extreme challenge of predicting around corners. For example, as shown in Fig. \ref{fig:teaser}, any future navigation trajectory passing the entrance of an unseen room requires the model to plausibly imagine the entire contents of that room (we dub this the \textit{room reveal} problem). This requires generalizing from the visual, spatial and semantic structure of previously explored environments---which in our case are photo-realistic 3D captures of real indoor spaces in the Matterport3D dataset~\cite{Matterport3D}. A third problem is temporal consistency:  predictions of unseen building regions should ideally be stochastic (capturing the full distribution of possible outcomes), but revisited regions should be rendered in a consistent manner to previous observations.

Towards this goal, we introduce \pathdreamer. Given one or more visual observations (consisting of RGB, depth and semantic segmentation for panoramas) of an indoor scene, \pathdreamer synthesizes high-resolution visual observations (RGB, depth and semantic segmentations) along a specified trajectory through future viewpoints, using a hierarchical two-stage approach. \pathdreamer's first stage, \segmodel, generates depth and semantic segmentations. Inspired by video prediction ~\cite{denton2018stochastic}, these outputs are conditioned on a latent noise tensor capturing the stochastic information about the next observation (such as the layout of an unseen room) that cannot be predicted deterministically. The second stage's \rgbmodel renders the depth and semantic segmentations as realistic RGB images using modified Multi-SPADE blocks~\cite{park2019semantic,mallya2020world}. To maintain long-term consistency in the generated observations, both stages use back-projected 3D point cloud representations which are re-projected into image space for context~\cite{mallya2020world}. 


\pathdreamer can generate plausible views for unseen indoor scenes under large viewpoint changes (see Figure \ref{fig:teaser}), while also addressing the \textit{room reveal} problem -- in this case correctly hypothesizing that the unseen room revealed at position 2 will most likely resemble a kitchen. Empirically, using the Matterport3D dataset~\cite{Matterport3D} and \threesixty observations, we evaluate both stages of our model against prior work and reasonable baselines and ablations. We find that the hierarchical structure of the model is essential for predicting over large viewpoint changes, that maintaining both RGB and semantic context is required, and that prediction quality degrades gradually when we evaluate with trajectory rollouts of up to 13m (with viewpoints 2.25m apart on average).

Encouraged by these results, we investigate whether \pathdreamer's RGB predictions can improve performance on a downstream task: Vision-and-Language Navigation (VLN), using the R2R dataset~\cite{anderson2018vision}. VLN requires agents to interpret and execute natural language navigation instructions in a photorealistic 3D environment. A robust finding from previous research is that performance improves dramatically when agents can look ahead at unobserved parts of the environment while following an instruction~\cite{majumdar2020improving}. We show that replacing look-ahead observations with \pathdreamer predictions maintains around half of the gains, a finding we expect to have significant implications for VLN research. In summary, our main contributions include:
\begin{compactitem}
    \item Proposing the study of visual world models for generic indoor environments and defining evaluation protocols and baselines for future work.
    \item \pathdreamer, a stochastic hierarchical visual world model combining multiple, independent threads of previous work on video prediction~\cite{denton2018stochastic}, semantic image synthesis~\cite{park2019semantic} and video-to-video synthesis~\cite{mallya2020world}.
    \item Extensive experiments characterizing the performance of \pathdreamer and demonstrating improved results on the downstream VLN task~\cite{anderson2018vision}.
\end{compactitem}

\cutsectionup
\section{Related Work}
\cutsectiondown

\paragraph{Video Prediction}

Our work is closely related to the task of video prediction, which aims to predict the future frames of a video sequence. While some video prediction methods predict RGB video frames directly~\cite{walker2014patch,aigner2018futuregan,kwon2019predicting,liu2020infinite}, many others use hierarchical models to first predict an intermediate representation (such as semantic segmentation)~\cite{luc2017predicting,jin2017predicting,wang2018video,xu2018structure,lee2021hierarchical}, which improves the fidelity of long-term predictions~\cite{lee2021hierarchical}. Several approaches have also incorporated 3D point cloud representations, using projective camera geometry to explicitly infer aspects of the next frame~\cite{vora2018future,mallya2020world,li2020street}. Inspired by this work, we adopt and combine both the hierarchical two-stage approach and 3D point cloud representations. 
Further, since our interest is in action-conditional world models, we provide a trajectory of future viewpoints to the model rather than assuming a constant frame rate and modeling camera motion implicitly, which is more typical in video generation~\cite{liu2020infinite,lee2021hierarchical}.




\cutparagraphup
\paragraph{Action-Conditional Video Prediction}
Conditional video prediction to improve agent reasoning and planning has been explored in several tasks. This includes video prediction for Atari games conditioned on control inputs~\cite{oh2015action, chiappa2017recurrent, osinski20, hafner2020mastering} and 3D game environments like Doom~\cite{ha2018world}. In robotics, action-conditional video prediction has been investigated for object pushing in tabletop settings to improve generalization to novel objects~\cite{finn2016unsupervised, finn2017deep, ebert2017self}. This work has been restricted to simple environments and low-resolution images, such as 64$\times$64 images of objects in a wooden box. To the best of our knowledge, we are the first to investigate action-conditional video prediction in building-scale environments with high-resolution (1024$\times$512) images.



\cutparagraphup
\paragraph{World Models and Navigation Priors}

World models~\cite{ha2018world} are an appealing way to summarize and distill knowledge about complex, high-dimensional environments. However, world models can differ in their outputs. While \pathdreamer predicts visual observations, there is also a vast literature on world models that predict compact latent representations of future states~\cite{karl2016deep,hafner2019dream,hafner2020mastering} or other task-specific measurements~\cite{dosovitskiy2016learning} or rewards~\cite{NIPS2017_ffbd6cbb}. This includes recent work attempting to learn statistical regularities and other priors for indoor navigation---for example, by mining spatial co-occurrences from real estate video tours~\cite{chang2020semantic}, learning to predict top-down belief maps over room characteristics~\cite{narasimhan2020seeing}, or learning to reconstruct house floor plans using audio and visual cues from a short video sequence~\cite{purushwalkam2020audio}. In contrast to these approaches, we focus on explicitly predicting visual observations (i.e., pixels) which are generic, human-interpretable, and apply to a wide variety of downstream tasks and applications. Further, recent work identifies a close correlation between image prediction accuracy and downstream task performance in model-based RL~\cite{babaeizadeh2020models}.

\cutparagraphup
\paragraph{Embodied Navigation Agents}
High-quality 3D environment datasets such as Matterport3D~\cite{Matterport3D}, StreetLearn~\cite{mirowski2019streetlearn,mehta2020retouchdown}, Gibson~\cite{xiazamirhe2018gibsonenv} and Replica~\cite{replica19arxiv} have triggered intense interest in developing embodied agents that act in realistic human environments~\cite{anderson2018evaluation}. Tasks of interest include ObjectNav~\cite{batra2020objectnav} (navigating to an instance of a particular kind of object), and Vision-and-Language Navigation (VLN)~\cite{anderson2018vision}, in which agents must navigate according to natural language instructions. Variations of VLN include indoor navigation~\cite{anderson2018vision,jain2019stay,Qi_2020_CVPR,ku2020room}, street-level navigation~\cite{chen2019touchdown,mehta2020retouchdown}, vision-and-dialog navigation~\cite{nguyen2019help,thomason2020vision,hahn2020way}, VLN in continuous environments~\cite{krantz2020navgraph}, and more. Notwithstanding considerable exploration of pretraining strategies~\cite{lu2019vilbert,hao2020towards,majumdar2020improving,zhu2020multimodal}, data augmentation approaches~\cite{fried2018speaker,fu2020counterfactual,tan2019learning}, agent architectures and loss functions~\cite{zhu2020auxiliary,ma2019self,ma2019regretful}, existing work in this space considers only model-free approaches. Our aim is to unlock model-based approaches to these tasks, using a visual world model to encode prior commonsense knowledge about human environments and thereby relieve the burden on the agent to learn these regularities. Underscoring the potential of this direction, we note that using the ground-truth environment for planning with beam search typically improves VLN success rates on the R2R dataset by 17-19\%~\cite{fried2018speaker,tan2019learning}.

\cutparagraphup
\paragraph{Novel View Synthesis}

We position our work with respect to novel view synthesis~\cite{flynn2016deep,kar2017learning,henzler2018single,flynn2019deepview,srinivasan2019pushing,zhou2018stereo,mildenhall2019local}. Methods for representing 3D scenes include point cloud representations~\cite{wiles2020synsin}, layered depth images~\cite{dhamo2019peeking}, and mesh representations~\cite{shih2020photography}. Recently, neural radiance fields (NeRF)~\cite{mildenhall2020nerf,martinbrualla2020nerfw,yu2020pixelnerf} achieved impressive results by capturing volume density and color implicitly with a neural network.  NeRF models synthesize very high quality 3D scenes, but a significant drawback for our purposes is that they require a large number of input views to render a single scene (e.g., 20--62 images per scene in \cite{mildenhall2020nerf}). More importantly, these models are typically trained to represent a single scene, and do not yet generalize well to unseen environments. In contrast, our problem demands generalization to unseen environments, using as little as one previous observation.

\begin{figure*}[t]
    \centering
    \includegraphics[width=1.0\textwidth]{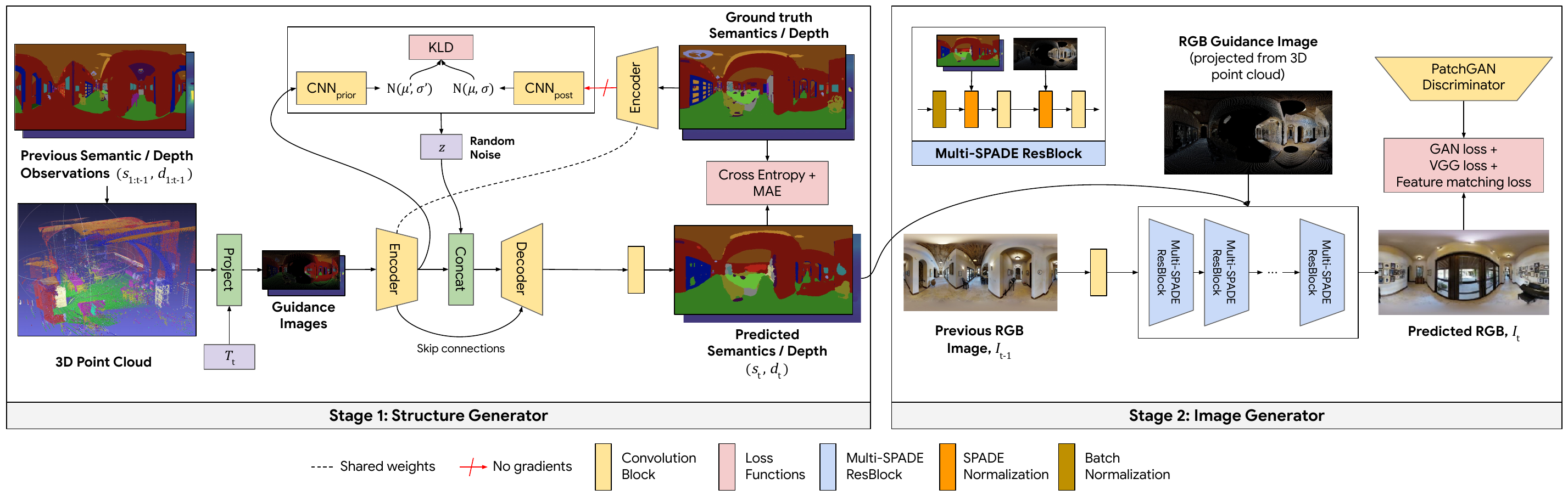}
    \vspace*{-0.25in}
    \caption{\pathdreamer model architecture at step $t$. Given a history of visual observations (RGB, depth and semantics) and a trajectory of future viewpoints, the \segmodel conditions on a sampled noise tensor before generating semantic and depth outputs to provide a high-level structural representation of the scene. Realistic RGB images are synthesized by the \rgbmodel in the second stage.}
    \vspace*{-0.2in}
    \label{fig:model}
\end{figure*} 

\cutsectionup
\section{\pathdreamer} \label{sec:approach}
\cutsectiondown


\pathdreamer is a world model that generates high-resolution visual observations from a trajectory of future viewpoints in buildings it has never observed. The input to \pathdreamer is a sequence of previous observations consisting of RGB images $I_{1:t-1}$, semantic segmentation images $s_{1:t-1}$, and depth images $d_{1:t-1}$ (where the depth and segmentations could be ground-truth or estimates from a model). We assume that a corresponding sequence of camera poses $T_{1:t-1}$ is available from an odometry system, and that the camera intrinsics are known or estimated. Our goal is to generate realistic RGB, semantic segmentation and depth images for a trajectory of future poses $T_{t}, T_{t+1}, \dots, T_{T}$, which may be provided up front or iteratively by some agent interacting with the returned observations. Note that we generate depth and segmentation because these modalities are useful in many downstream tasks. We assume that the future trajectory may traverse unseen areas of the environment, requiring the model to not only in-fill minor object dis-occlusions, but also to imagine entire room reveals (Figure \ref{fig:teaser}).  


%


Figure \ref{fig:model} shows our proposed hierarchical two-stage model for addressing this challenge. It uses a latent noise tensor $z_{t}$ to capture the stochastic information about the next observation (e.g. the layout of an unseen room) that cannot be predicted deterministically. Given a sampled noise tensor $z_{t}$, the first stage (\segmodel) generates a new depth image $\hat{d}_{t}$ and segmentation image $\hat{s}_{t}$ to provide a plausible high-level semantic representation of the scene, using as context the previous semantic and depth images $s_{1:t-1}$, $d_{1:t-1}$. In the second stage (\rgbmodel), the predicted semantic and depth images $\hat{s}_{t}$, $\hat{d}_{t}$ are rendered into a realistic RGB image $\hat{I}_{t}$ using previous RGB images $I_{1:t-1}$ as context. In each stage, context is provided by accumulating previous observations as a 3D point cloud which is re-projected into 2D using $T_{t}$. 




\cutsubsectionup
\subsection{\segmodel: Segmentation \& Depth} \label{sec:semanticapproach}
\cutsubsectiondown

\pathdreamer's first stage is the \segmodel, a stochastic encoder-decoder network for generating diverse, plausible segmentation and depth images. Like \cite{mallya2020world}, to provide the previous observation context, we first back-project the previous segmentations $s_{1:t-1}$ into a unified 3D semantic point cloud using the depth images $d_{1:t-1}$ and camera poses $T_{1:t-1}$. We then re-project this point cloud back into pixel space using $T_{t}$ to create sparse segmentation and depth guidance images $s^{'}_{t}$, $d^{'}_{t}$ which reflect the current pose.


The input to the encoder is a one-hot encoding of the semantic guidance image $s^{'}_{t} \in \mathbb{R}^{W \times H \times C}$, concatenated with the depth guidance image $d^{'}_{t} \in \mathbb{R}^{W \times H \times 1}$. The architecture of the encoder-decoder model is based on RedNet~\cite{jiang2018rednet} -- a ResNet-50~\cite{he2016deep} architecture designed for indoor RGB-D semantic segmentation. RedNet uses transposed convolutions for upsampling in the decoder and skip connections between the encoder and decoder to preserve spatial information. Since the input contains a segmentation image, and segmentation classes differ across datasets, the encoder-decoder is not pretrained. We introduce the latent spatial noise tensor $z_t \in \mathbb{R}^{H' \times W' \times 32}$ into the model by concatenating it with the feature map between the encoder and the decoder. The final output of the encoder-decoder model is a segmentation image $\hat{s}_{t}$ and a depth image $\hat{d}_{t}$, with segmentation predictions generated by a $C$-way softmax and depth outputs normalized in the range $(0, 1)$ and generated via a sigmoid function. At each step during inference, the segmentation prediction $\hat{s}_{t}$ is back-projected and added to the point cloud to assist prediction in future timesteps.



\begin{figure*}[t]
    \centering
    \includegraphics[width=1.0\linewidth]{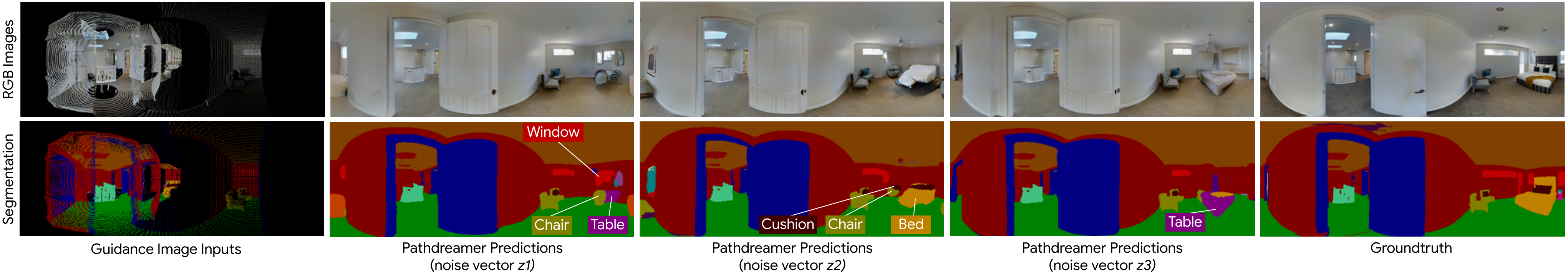}\\
    \includegraphics[width=1.0\linewidth]{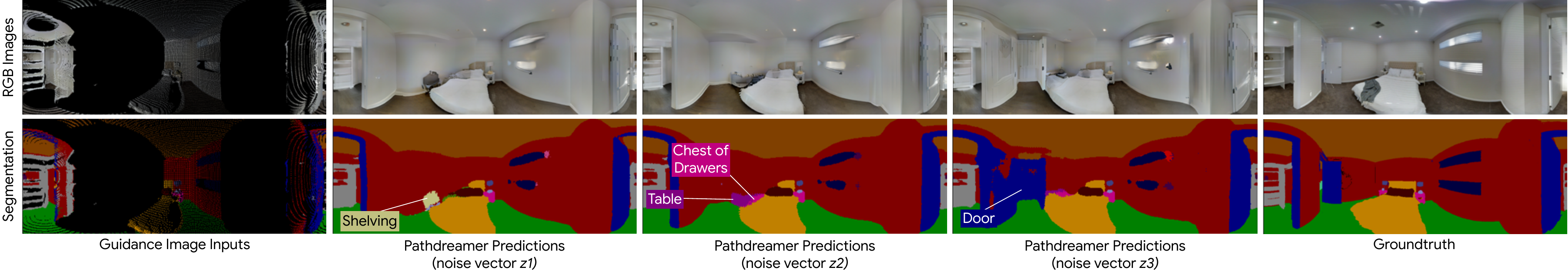}
    \vspace{-0.3in}
     \caption{When predicting around corners, the \segmodel can sample diverse and semantically plausible scene layouts which are closely reflected in the RGB output of the \rgbmodel, shown here for two guidance image inputs (left columns; unseen areas are indicated by solid black regions). Each example shows three alternative \textit{room reveals} and the groundtruth. In the bottom example, the model considers various completions for a bedroom but fails to anticipate the groundtruth's matching lamp on the opposite side of the bed.
    }
    \vspace{-0.2in}
    \label{fig:noise_qualitative}
\end{figure*}

To generate the noise tensor $z_{t}$, we take inspiration from SVG~\cite{denton2018stochastic} and learn a conditional prior noise distribution $p_{\psi}(z_t | s^{'}_t, d^{'}_t)$. Intuitively, there are many possible scenes that may be generated for an unseen building region. We would like $z_t$ to carry the stochastic information about the next observation that the deterministic encoder cannot capture, and we would like for the decoder to make good use of that information. During training, we encourage the first outcome by using a KL-divergence loss to force the prior distribution $p_{\psi}(z_t | s^{'}_t, d^{'}_t)$ to be close to the posterior distribution ${\phi}(z_t | s_t, d_t)$ which is conditioned on the ground-truth segmentation and depth images. We encourage the second outcome by providing the decoder with sampled $z_t$ values from the posterior distribution $q_{\phi}$ (conditioned on the ground-truth outputs) during training. During inference, the latent noise $z_t$ is sampled from the prior distribution $p_{\psi}$ and the posterior distribution $q_{\phi}$ is not used. Both distributions are modeled using 3-layer CNNs that take their input from the encoder and output two channels representing $\mu$ and $\sigma$ to parameterize a multivariate Gaussian distribution $\mathcal{N}(\mu, \sigma)$. As shown in Figure~\ref{fig:noise_qualitative}, the noise is useful in encoding diverse, plausible representations of unseen regions.


Overall, the \segmodel is trained to minimize a joint loss consisting of a cross-entropy loss $L_{\text{ce}}$ for semantic predictions, a mean absolute error term for depth predictions, and the KL-divergence term for the noise tensor:
\begin{align}
\mathcal{L}_{\text{Structure}} &= \lambda_{\text{ce}} L_{\text{ce}} (s_t, \hat{s}_t) \nonumber\\
  &+ \lambda_{\text{d}} \norm{d_t - \hat{d}_t}_{1} \nonumber\\
  &+ \lambda_{\text{KL}} D_{\text{KL}} \big(q_{\phi}(z_t | s_t, d_t), p_{\psi}(z_t | s^{'}_t, d^{'}_t) \big)
\end{align}

\noindent
where $\lambda_{\text{ce}}$, $\lambda_{\text{d}}$, and $\lambda_{\text{KL}}$ are weights determined by a grid search. We set these to 1, 100, and 0.5 respectively.

\cutsubsectionup
\subsection{\rgbmodel: RGB}
\cutsubsectiondown


The \rgbmodel is an image-to-image translation GAN~\cite{goodfellow2014generative,wang2018high} that converts the semantic and depth predictions $\hat{s}_t$, $\hat{d}_t$ from the first stage into a realistic RGB image $\hat{I}_t$. Our model architecture is based on SPADE blocks~\cite{park2019semantic} that use spatially-adaptive normalization layers to insert context into multiple layers of the network. As with our \segmodel, we maintain an accumulating 3D point cloud containing all previous image observations. This provides a sparse RGB guidance image $I'_t$ when re-projected. Similar to Multi-SPADE~\cite{mallya2020world}, we insert two SPADE normalization layers into each residual block: one conditioned on the concatenated semantic and depth inputs $[\hat{s}_t, \hat{d}_t]$, and one conditioned on the RGB guidance image $I'_t$. The sparsity of the RGB guidance image is handled by applying partial convolutions~\cite{liu2018image}. In total \rgbmodel consists of 7 Multi-SPADE blocks, preceded by a single convolution block.


Following SPADE~\cite{park2019semantic}, the model is trained with the GAN hinge loss, feature matching loss~\cite{wang2018high}, and perceptual loss~\cite{johnson2016perceptual} from a pretrained VGG-19~\cite{simonyan2014very} model. During training, the generator is provided with the ground-truth segmentation image $s_t$ and ground-truth depth image $d_t$. Our discriminator architecture is based on PatchGAN~\cite{pix2pix2017}, and takes as input the concatenation of the ground-truth image $I_t$ or generated image $\hat{I}_t$, the ground-truth depth image $d_t$ and the ground-truth semantic image $s_t$. The losses for the generator $G$ and the discriminator $D$ are:
\begin{align}
\mathcal{L}_{G} &= -\lambda_{\text{GAN}} \mathbb{E}_{x_t}\left[D(G(x_t))\right] \nonumber\\
  &+ \lambda_{\text{VGG}} \sum_{i=1}^n \frac{1}{n} \norm{\phi^{(i)}(I_t) - \phi^{(i)}(G(x_t))}_1 \nonumber\\
  &+ \lambda_{\text{FM}} \sum_{i}^n \frac{1}{n} \norm{D^{(i)}(I_t) - D^{(i)}(G(x_t))}_1 \\
\mathcal{L}_D=&-\mathbb{E}_{x_t}\left[\min(0, -1 + D(I_t))\right] \nonumber\\
&- \mathbb{E}_{x_t}\left[\min(0, -1- D(G(x_t)))\right]
\end{align}
where $x_t = (s_t, d_t, I'_t)$ denotes the complete set of inputs to the generator, $\phi^{(i)}$ denotes the output of the $i^{\text{th}}$ layer of the pretrained VGG-19 model, $D^{(i)}$ denotes the output of the discriminator's $i$-th layer (conditioning inputs $s_t, d_t$ to the discriminator have been dropped to save space). Like the \segmodel, the \rgbmodel is not pretrained.

\begin{figure*}[t]
    \centering
    \includegraphics[width=1.0\textwidth]{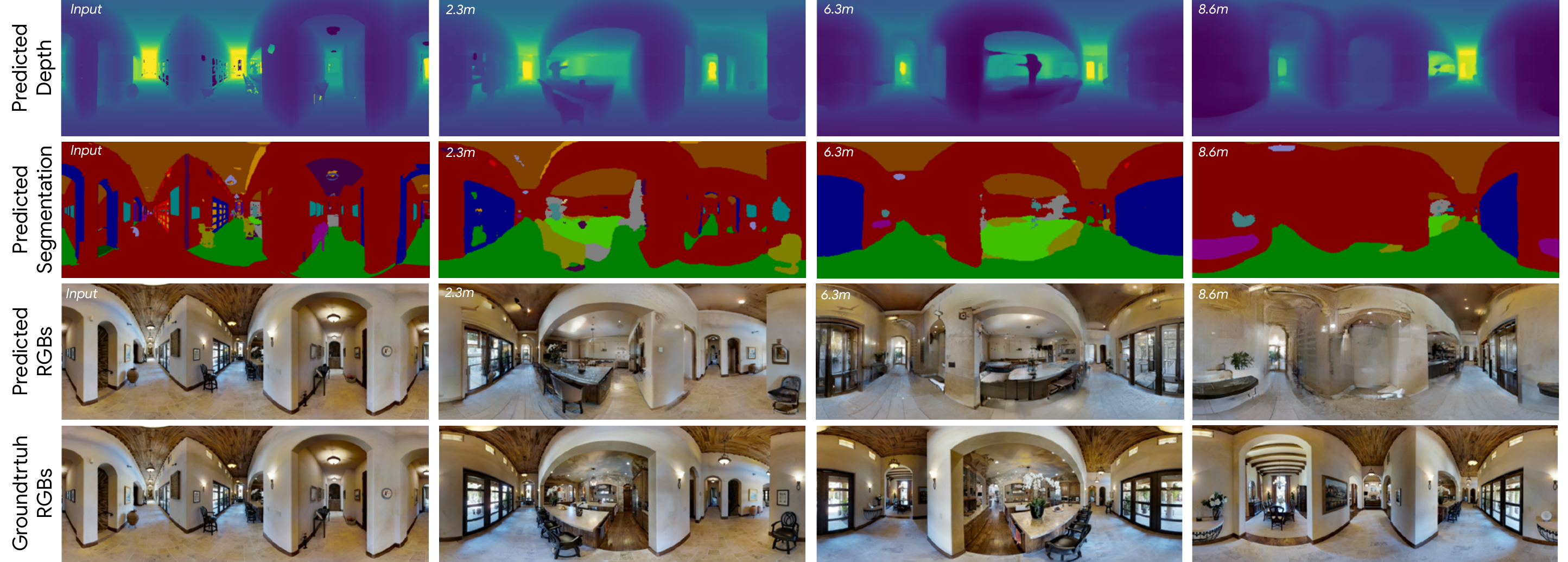}
    \vspace{-20pt}
    \caption{Example full prediction sequence beginning with one observation (depth, semantics, RGB) as context and generating observations for 3 new viewpoints traversing a corridor. At 2.3m the model completes a room reveal, imagining a kitchen-like space. After 8.6m the model's predictions degrade. More examples are provided in the supplementary.}
    \vspace{-10pt}
    \label{fig:qualitative}
\end{figure*}

\cutsubsectionup
\subsection{Training and Inference}
\cutsubsectiondown
\label{sec:training}

\paragraph{Dataset} For training and evaluation we use Matterport3D~\cite{Matterport3D}, a dataset of 10.8k RGB-D images from 90 building-scale indoor environments. For each environment, Matterport3D also includes a textured 3D mesh which is annotated with 40 semantic classes of objects and building components. To align with downstream VLN tasks, in all experiments the RGB, depth and semantic images are \threesixty panoramas in equirectangular format. 

\cutparagraphup
\paragraph{Trajectories} To train \pathdreamer, we sampled 400k trajectories from the Matterport3D training environments. To define feasible trajectories, we used the navigation graphs from the Room-to-Room (R2R) dataset~\cite{anderson2018vision}, in which nodes correspond to panoramic image locations, and edges define navigable state transitions. For each trajectory 5--8 panoramas were sampled, choosing the starting node and the edge transitions uniformly at random. On average the viewpoints in these trajectories are 2m apart. Training with relatively large viewpoint changes is desirable, since the model learns to synthesize observations with large viewpoint changes in a single step (without the need to incur the computational cost of generating intervening frames). However, this does not preclude \pathdreamer from generating smooth video outputs at high frame rates\footnote{See \url{https://youtu.be/StklIENGqs0} for our video generation results.}.   

\cutparagraphup
\paragraph{Training} The first and second stages of the model are trained separately. For the \rgbmodel, we use the Matterport3D RGB panoramas as training targets at 1024$\times$512 resolution. We use the Habitat simulator~\cite{habitat19iccv} to render ground-truth depth and semantic training inputs and stitch these into equirectangular panoramas. We perform data augmentation by randomly cropping and horizontally rolling the RGB panoramas, which we found essential due to the limited number of panoramas available.

To train the \segmodel, we again used Habitat to render depth and semantic images. Since this stage does not require aligned RGB images for training, in this case we performed data augmentation by perturbing the viewpoint coordinates with a random Gaussian noise vector drawn from $\mathcal{N}(0, 0.2\text{m})$ independently along each 3D axis. The \segmodel was trained with equirectangular panoramas at 512$\times$256 resolution.

\cutparagraphup
\paragraph{Inference} 
To avoid heading discontinuities during inference, we use circular padding on the image x-axis for both the \segmodel and the \rgbmodel. The 512$\times$256 resolution semantic and depth outputs of the \segmodel are upsampled to 1024$\times$512 using nearest neighbor interpolation before they are passed to the \rgbmodel. In quantitative experiments, we set the \segmodel noise tensor $z_t$ to the mean of the prior.

\cutsectionup
\section{Experiments} \label{sec:results}
\cutsectiondown

For evaluation we use the paths from the \valseen and \valunseen splits of the R2R dataset~\cite{anderson2018vision}. \valseen contains 340 trajectories from environments in the Matterport3D training split. \valunseen contains 783 trajectories in Matterport3D environments not seen in training. Since R2R trajectories contain 5-7 panoramas and at least 1 previous observation is given as context, we report evaluations over 1--6 steps, representing predictions over trajectory rollouts of around 2--13m (panoramas are 2.25m apart on average). See Figure \ref{fig:qualitative} for an example rollout over 8.6m. We characterize the performance of \pathdreamer in comparison to baselines, ablations and in the context of the downstream task of Vision-and-Language Navigation (VLN).


\cutsubsectionup
\subsection{\pathdreamer Results}
\cutsubsectiondown

\paragraph{Semantic Generation} A key feature of our approach is the ability to generate semantic segmentation and depth outputs, in addition to RGB.
We evaluate the generated semantic segmentation images using mean Intersection-Over-Union (mIOU) and report results for:
\begin{compactitem}
    \item \textbf{Nearest Neighbor}: A baseline without any learned components, using nearest-neighbor interpolation to fill holes in the projected semantic guidance image $s'_t$.
    \item \textbf{Ours (Teacher Forcing)}: \segmodel trained using the ground truth semantic and depth images as the previous observation at every time step.
    \item \textbf{Ours (Recurrent)}: \segmodel trained while feeding back its own semantic and depth predictions as previous observations for the next step prediction. This reduces train-test mismatch and may allow the model to compensate for errors when doing longer roll-outs.
\end{compactitem}

\noindent
We also tried training the hierarchical convolutional LSTM from \cite{lee2021hierarchical}, but found that it frequently collapsed to a single class prediction. We attribute this to the large viewpoint changes and heavy occlusion in the training sequences; we believe this can be more effectively modeled with point cloud geometry than with a geometry-unaware LSTM. 



\begin{figure}[t]
\begin{center}
\begin{subfigure}{.49\textwidth}
\centering
\includegraphics[width=0.49\linewidth]{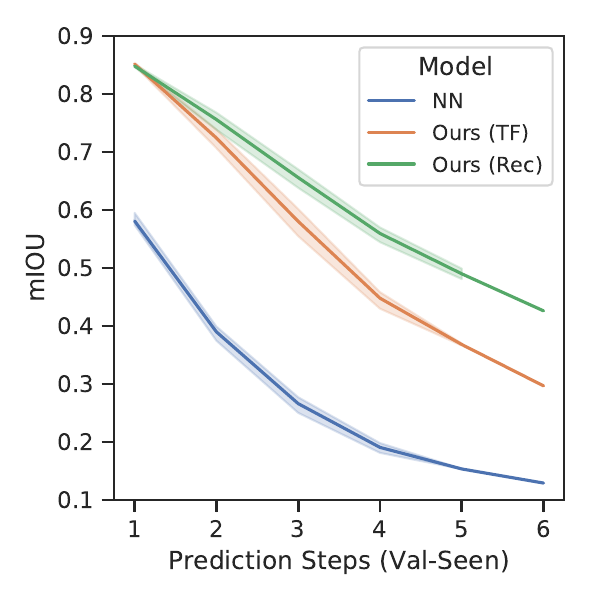}
\includegraphics[width=0.49\linewidth]{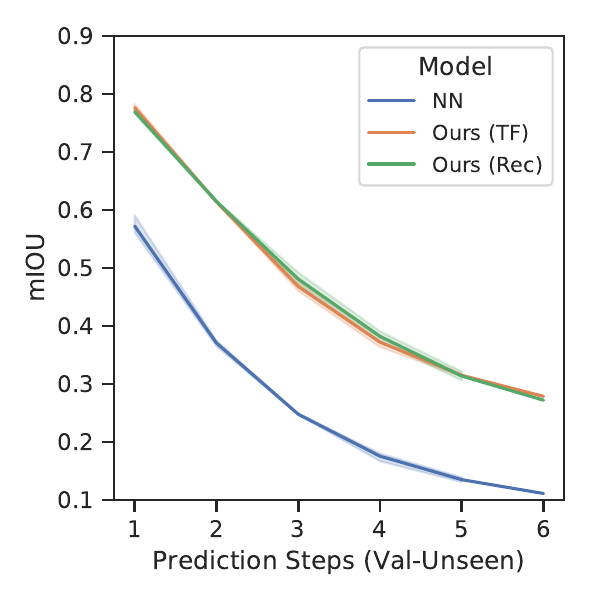}
\vspace{-7pt}
\caption{Semantic segmentation mean-IOU ($\uparrow$). \textbf{[TF]:}~Teacher Forcing. \textbf{[Rec]:}~Recurrent.}\label{fig:miou-plot}
\end{subfigure}
\begin{subfigure}{.49\textwidth}
\centering
\includegraphics[width=0.49\linewidth]{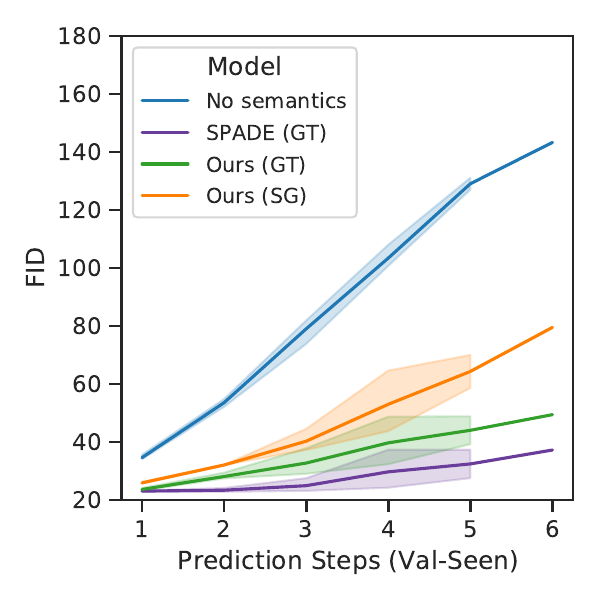}
\includegraphics[width=0.49\linewidth]{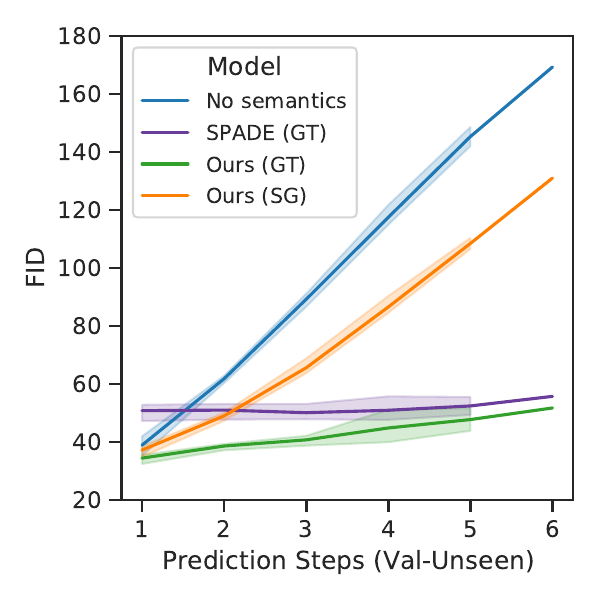}
\vspace{-7pt}
\caption{RGB generation FID ($\downarrow$). \textbf{[GT]}:~Ground truth semantic inputs. \textbf{[SG]}:~Structure Generator predictions.}\label{fig:fid-plot}
\end{subfigure}
\end{center}
\vspace{-15pt}
\caption{\pathdreamer semantic segmentation mean-IOU (above) and RGB generation FID (below). Results are shown for \valseen (left) and \valunseen (right). Confidence intervals indicate the range of outcomes with 1, 2 or 3 previous observations as context.}
\vspace{-0.2in}
\label{fig:plots}
\end{figure}

As illustrated in Table \ref{table:segmentation_results}, \pathdreamer performs far better than the Nearest Neighbor baseline regardless of the number of steps in the rollout or the number of previous observations used as context. As expected, performance in seen environments is higher than unseen. Perhaps surprisingly, in Figure \ref{fig:miou-plot} we show that Recurrent training improves results during longer rollouts in the training environments (\valseen), but this does not improve results on \valunseen, perhaps indicating that the error compensation learned by the \rgbmodel does not easily generalize.

In addition to accurate predictions, we also want generated results to be \textit{diverse}. Figure~\ref{fig:noise_qualitative} shows that our model can generate diverse semantic scenes by interpolating the noise tensor $z_t$, and that the RGB outputs closely reflect the generated semantic image. This allows us to generate multiple plausible alternatives for the same navigation trajectory.

\cutparagraphup
\paragraph{RGB Generation}
To evaluate the quality of RGB panoramas generated by the \rgbmodel, we compute the Fréchet Inception Distance (FID) \cite{heusel2017gans} between generated and real images for each step in the paths. We report results using the semantic images generated by the \segmodel as inputs (i.e., our full model). To quantify the potential for uplift with better \segmodel{s}, we also report results using ground truth semantic segmentations as input. We compare to two ablated versions of our model:
\begin{compactitem}
    \item \textbf{No Semantics}: The semantic and depth inputs $s_t$, $d_t$ are removed from the Multi-SPADE blocks.
    \item \textbf{SPADE}: An ablation of the RGB inputs to the model, comprising the previous RGB image $I_{t-1}$ and the re-projected RGB guidance image $I'_t$. The semantic image $s_t$ replaces $I_{t-1}$ as input to the model and the $I'_t$ input layers are removed from the Multi-SPADE blocks, making this effectively the SPADE model~\cite{park2019semantic}. 
\end{compactitem}


\begin{table}[t]
\begin{center}
\footnotesize
\setlength\tabcolsep{2pt}
\begin{tabularx}{\linewidth}{lcccccc}
 &  & \multicolumn{2}{c}{\textbf{\valseen}} & & \multicolumn{2}{c}{\textbf{\valunseen}} \\
\cmidrule{3-4}\cmidrule{6-7}
\textbf{Model} & \textbf{Context} & 1 Step & 1--6 Steps & & 1 Step & 1--6 Steps \\ \midrule
Nearest Neighbor & 1 & 59.5 & 32.0 & & 59.1 & 30.6  \\
Ours (Teacher Forcing) & 1 & \textbf{84.9} & 59.2 & & \textbf{78.3} & 50.8 \\
Ours (Recurrent) & 1 & 84.7 & \textbf{65.9} & &  77.5 & \textbf{50.9} \\
\midrule
Nearest Neighbor & 2 & 57.4 & 35.2 & &  56.5 & 33.8  \\
Ours (Teacher Forcing) & 2 & \textbf{85.4} & 64.6 & &  \textbf{77.4} & 55.5 \\
Ours (Recurrent) & 2 & 85.1 & \textbf{70.2} & &  76.6 & \textbf{55.7} \\
\midrule
Nearest Neighbor  & 3 & 57.4 & 38.7 & &  56.1 & 37.7 \\
Ours (Teacher Forcing) & 3 & \textbf{85.1} & 68.5 & &  \textbf{77.3} & 60.4 \\
Ours (Recurrent) & 3 & 84.6 & \textbf{72.7} & &  76.8 & \textbf{60.8} \\
\bottomrule
\end{tabularx}
\end{center}
\vspace{-15pt}
\caption{Mean-IOU ($\uparrow$) for generated semantic segmentations with varying context and prediction steps. }
\vspace{-5pt}
\label{table:segmentation_results}
\end{table}

\begin{table}[t]
\begin{center}
\footnotesize
\setlength\tabcolsep{1pt}
\begin{tabularx}{\linewidth}{lcccccccccc}
 &  & \multicolumn{3}{c}{\textbf{Inputs}} & & \multicolumn{2}{c}{\textbf{\valseen}} & & \multicolumn{2}{c}{\textbf{\valunseen}} \\
\cmidrule{3-5} \cmidrule{7-8} \cmidrule{10-11}
\textbf{Model} & \textbf{Context} & Obs & Sem & RGB & & 1 Step & 1--6 Steps & & 1 Step & 1--6 Steps \\ \midrule
No Semantics & 1 & - & & \checkmark & & 34.1 & 81.7 & & 35.2 & 90.8  \\
SPADE & 1 & GT & \checkmark & & & \textbf{23.3} & \textbf{24.9} &  & 47.3 & 50.3  \\
Ours & 1 & GT & \checkmark & \checkmark & & 24.1 & 31.3 & & \textbf{32.4} & \textbf{39.9} \\
Ours & 1 & SG & \checkmark & \checkmark & & 26.2 & 41.7 & & 34.8 & 70.4 \\
\midrule
No Semantics & 2 & - & & \checkmark & & 34.0 & 69.0 & & 39.5 & 78.6  \\
SPADE & 2 & GT & \checkmark & & & \textbf{22.8} & \textbf{25.3} & & 52.3 & 51.2  \\
Ours & 2 & GT & \checkmark & \checkmark & & 23.5 & 31.2 & & \textbf{35.3} & \textbf{39.9} \\
Ours & 2 & SG & \checkmark & \checkmark & & 25.8 & 38.4 &  & 38.2 &	61.0 \\
\midrule
No Semantics & 3 & - & & \checkmark & & 35.6 & 60.4 & & 41.9 & 67.5 \\
SPADE & 3 & GT & \checkmark & & & \textbf{23.1} & \textbf{26.2} & & 52.8 & 50.7  \\
Ours & 3 & GT & \checkmark & \checkmark & & 23.2 & 31.7 & & \textbf{35.5} & \textbf{39.2} \\
Ours & 3 & SG & \checkmark & \checkmark & & 25.6 & 36.7 & & 38.5 & 52.9 \\
\bottomrule
\end{tabularx}
\end{center}
\vspace{-10pt}
\caption{FID scores ($\downarrow$) for generated RGB images with varying context and prediction steps, using either ground truth semantics (GT) or \segmodel predictions (SG) as input.}
\vspace{-10pt}
\label{table:rgb_results}
\end{table}  

\begin{figure}[t]
\begin{center}
\includegraphics[width=1.0\linewidth]{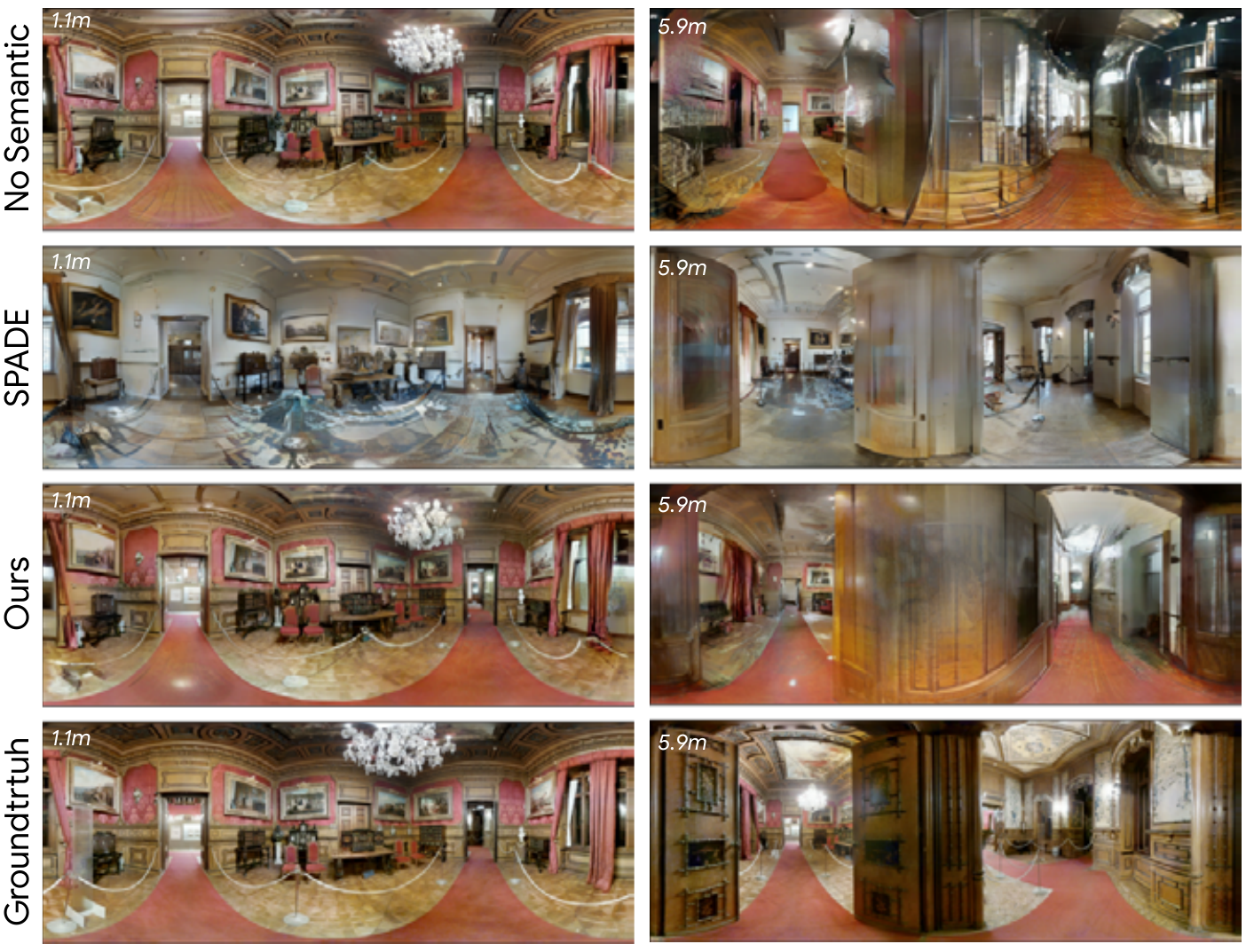}
\vspace{-0.7cm}
\end{center}
\vspace{-0.12in}
\caption{Visual comparison of ablated \rgbmodel outputs on \valunseen with \textit{ground truth} segmentation and depth. Both RGB and semantic context are important for best performance.}
\vspace{-0.2in}
\label{fig:ablation-qualitative}
\end{figure}

As shown in Table~\ref{table:rgb_results}, SPADE performs the best in \valseen, indicating that the model has the capacity to memorize the training environments. In this case, RGB inputs are not necessary. However, our model performs noticeably better in \valunseen, highlighting the importance of maintaining RGB context in unseen environments (which is our focus). Performance degrades significantly in the No Semantics setting in both \valseen and \valunseen. We observed that without semantic inputs, the model is unable to generate meaningful images over longer horizons, which validates our two-stage hierarchical approach. These results are reflected in the FID scores, as well as qualitatively (Figure~\ref{fig:ablation-qualitative}); \rgbmodel's outputs are significantly crisper, especially over longer horizons. Due to the benefit of guidance images, the \rgbmodel's textures are also generally better matched with the unseen environment, while SPADE tends to wash out textures, usually creating images of a standard style. 
Figure \ref{fig:fid-plot} plots performance for every setting step-by-step. FID of the \rgbmodel improves substantially when using ground truth semantics, particularly for longer rollouts, highlighting the potential to benefit from improvements to the \segmodel.

\cutsubsectionup
\subsection{VLN Results}
\cutsubsectiondown

Finally, we evaluate whether predictions from \pathdreamer can improve performance on a downstream visual navigation task. We focus on Vision-and-Language Navigation (VLN) using the R2R dataset~\cite{anderson2018vision}. Because reaching the navigation goal requires successfully grounding natural language instructions to visual observations, this provides a challenging task-based assessment of prediction quality.

In our inference setting, at each step while moving through the environment we use a baseline VLN agent based on \cite{wang2019reinforced} to generate a large number of possible future trajectories using beam search. We then rank these alternative trajectories using an instruction-trajectory compatibility model~\cite{discriminator21} to assess which trajectory best matches the instruction. The agent then executes the first action from the top-ranked trajectory before repeating the process. We consider three different planning horizons, with future trajectories containing 1, 2 or 3 forward steps. 

The instruction-trajectory compatibility model is a dual-encoder that separately encodes textual instructions and trajectories (encoded using visual observations and path geometry) into a shared latent space. To improve performance on incomplete paths, we introduce truncated paths into the original contrastive training scheme proposed in \cite{discriminator21}. The compatibility model is trained using only ground truth observations. However, during inference, RGB observations for future steps are drawn from three different sources:

\begin{compactitem}
    \item \textbf{Ground truth}: RGB observations from the actual environment, i.e., look-ahead observations.
    \item \textbf{Pathdreamer}: RGB predictions from our model.
    \item \textbf{Repeated pano}: A simple baseline in which the most recent RGB observation is repeated in future steps.
    \item \textbf{Blank pano}: A simple baseline in which blank images are provided as future observations. 
\end{compactitem}
\noindent
Note that in all cases the geometry of the future trajectories is determined by the ground truth R2R navigation graphs. In Table \ref{table:vln_results}, we report \valunseen results for this experiment using standard metrics for VLN: navigation error (NE), success rate (SR), shortest path length (SPL), normalized Dynamic Time Warping (nDTW)~\cite{ilharco2019general}, and success weighted by normalized Dynamic Time Warping (sDTW)~\cite{ilharco2019general}.

Consistent with prior work~\cite{fried2018speaker,tan2019learning}, we find that looking ahead using ground truth visual observations provides a robust performance boost, e.g., success rate increases from 44.6\% with 1 planning step (top panel) to 59.3\% with 3 planning steps (bottom panel). At the other extreme, the Repeated pano baseline is weak, with a success rate of just 35.7\% with 1 planning step (top row). The Blank pano baseline is similar, with a success rate of 35.9\%. This is not surprising: these baselines deny the compatibility model any useful visual representation of the next action, which is crucial to performance~\cite{fried2018speaker,tan2019learning}. However, increasing the planning horizon does improve performance even for the Repeated/Blank pano baselines, since the compatibility model is able to compare the geometry of alternative future trajectories. Finally, we observe that using \pathdreamer's visual observations closes about half the gap between the Repeated pano baseline and the ground truth observations, e.g., 50.4\% success with \pathdreamer vs. 40.6\% and 59.3\% respectively for the others. We conclude that using \pathdreamer as a visual world model can improve performance on downstream tasks, although existing agents still rely on using a navigation graph to define the feasible action space at each step. \pathdreamer is complementary to current SOTA model-based approaches, and a combination would likely lead to further boosts in VLN performance, which is worth investigating in future work.


\begin{table}[]
\begin{center}
\footnotesize
\setlength\tabcolsep{3pt}
\begin{tabularx}{\linewidth}{lcccccc}
\textbf{Observations} & \textbf{Plan Steps} &   \textbf{NE} $\downarrow$   &  \textbf{SR} $\uparrow$    & \textbf{SPL} $\uparrow$   & \textbf{nDTW} $\uparrow$ & \textbf{sDTW} $\uparrow$   \\
\midrule
Repeated pano         & 1                &  6.75    &   35.7   &  33.8   &   52.0    &   31.2  \\
Blank pano         & 1                &  7.29  &   35.9   &  33.7   &   50.9 &   31.5  \\
Pathdreamer             & 1                &  6.55    &   39.9   &  38.3   &   54.6    &   35.2 \\
Ground truth          & 1                &  5.80    &   44.6   &  42.7   &   58.9    &   39.4    \\
\midrule
Repeated pano         & 2                &  6.76     &  36.8    &  34.0   &    51.8   &   31.7       \\
Blank pano         & 2                &  6.65     &  40.0    &  37.2   &    53.3   &   34.8       \\
Pathdreamer             & 2                &  5.8     &  46.5    &  43.9   &    59.1   &   41.2   \\
Ground truth          & 2                &  4.95     &  54.3    &  51.3   &    64.9   &   48.3 \\
\midrule
Repeated pano         & 3                &  6.25    &  40.6    &  37.7   &    55.6   &   35.2  \\
Blank pano         & 3                &  6.48    &  41.9    &  38.8   &    54.2   &   36.1  \\
Pathdreamer             & 3                &  5.32    &  50.4    &  47.3   &    61.8   &   44.4  \\
Ground truth          & 3                &  4.44    &  59.3    &  55.8   &    67.9   &   52.7  \\
\bottomrule
\end{tabularx}
\end{center}
\vspace{-0.2in}
\caption{VLN \valunseen results using an instruction-trajectory compatibility model to rank alternative future trajectories with planning horizons of 1, 2 or 3 steps.}
\label{table:vln_results}
\vspace{-0.15in}
\end{table}

\cutsectionup
\section{Conclusion}
\cutsectiondown

\pathdreamer is a stochastic hierarchical visual world model that can synthesize realistic and diverse \threesixty panoramic images for unseen trajectories in real buildings.
As a visual world model, \pathdreamer also shows strong promise in improving performance on downstream tasks, which we show with VLN. Most notably, we show that \pathdreamer captures around half the benefit of looking ahead at actual observations from the environment. The efficacy of \pathdreamer in the VLN task may be attributed to its ability to model fundamental constraints in the real world, and thus relieve agents from having to learn the geometry and visual and semantic structure of buildings. Applying \pathdreamer to other embodied navigation tasks such as Object-Nav~\cite{batra2020objectnav}, VLN-CE~\cite{krantz2020navgraph} and street-level navigation~\cite{chen2019touchdown,mehta2020retouchdown} are natural directions for future work.


{\small
\bibliographystyle{ieee_fullname}
\bibliography{references}
}

\pagebreak

\appendix
\section{Qualitative Results}
\begin{figure*}[t]
\centering
    \begin{subfigure}{0.97\textwidth}
    \includegraphics[width=1.0\linewidth]{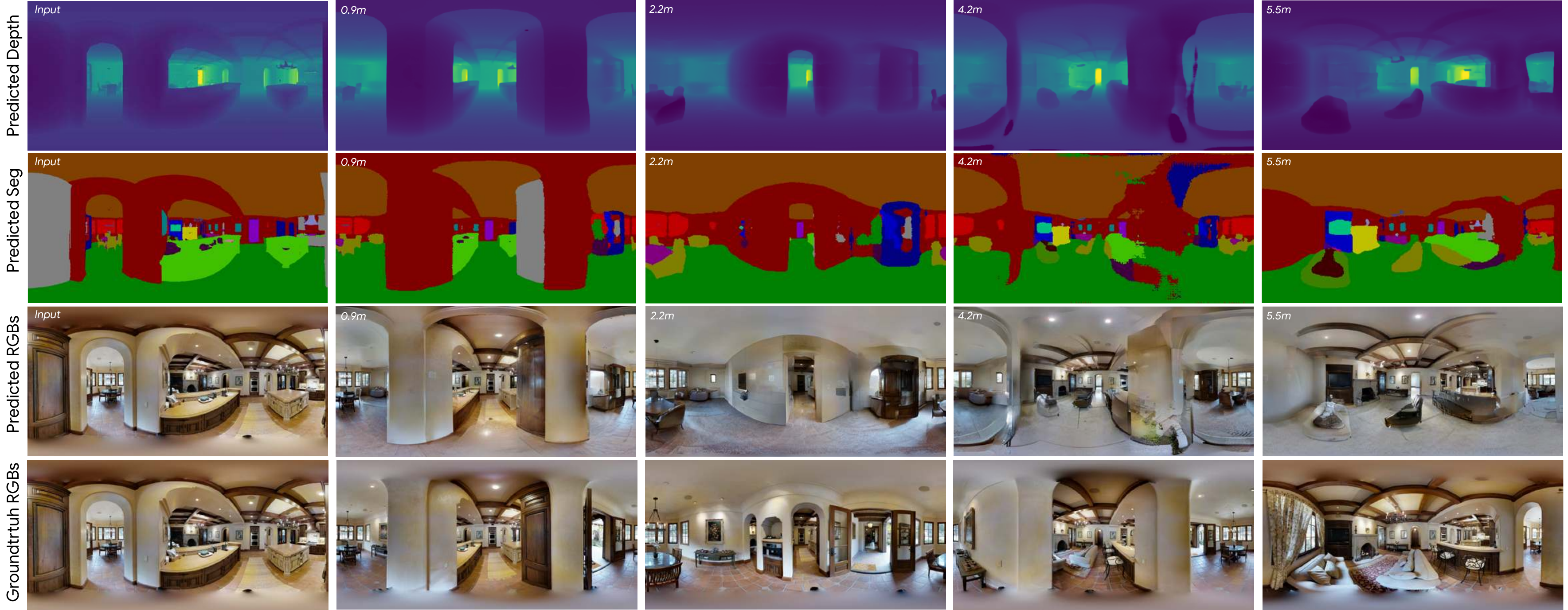}
    \vspace{-17pt}
    \caption{At 5.5m, \pathdreamer completes a room reveal -- imagining a living room space. Despite differences in details (e.g. door on the left rather than a window, and placement of a couch closer to the right), the result is quite close to the ground truth.}
    \vspace{3pt}
    \label{fig:qualitative_54}
    \end{subfigure}

    \begin{subfigure}{0.97\textwidth}
    \includegraphics[width=1.0\linewidth]{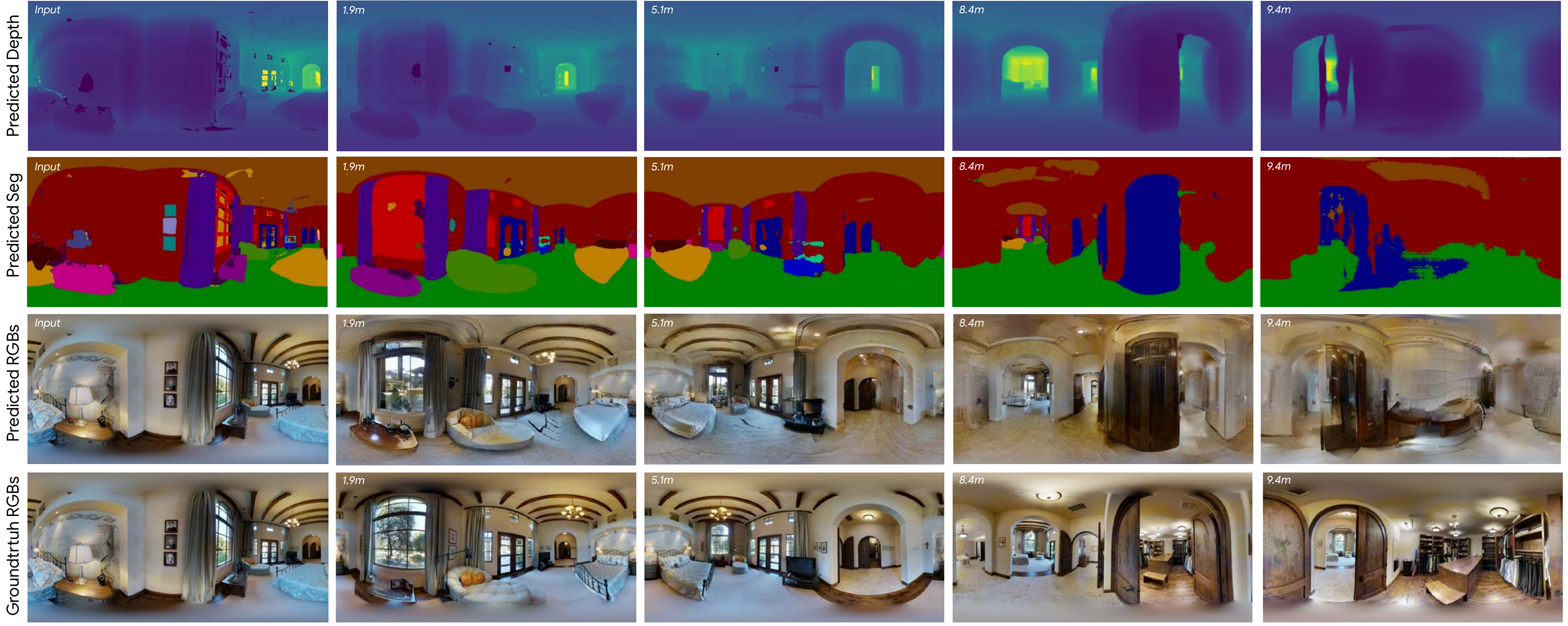}
    \vspace{-17pt}
    \caption{In this example, the navigation sequence exits a bedroom. \pathdreamer is able to maintain consistency and realism for unseen corners of the room (1.9m and 5.1m). At 8.4m, \pathdreamer also generates a plausible hallway seen after exiting the room.}
    \vspace{3pt}
    \label{fig:qualitative_190}
    \end{subfigure}

    \begin{subfigure}{0.97\textwidth}
    \includegraphics[width=1.0\linewidth]{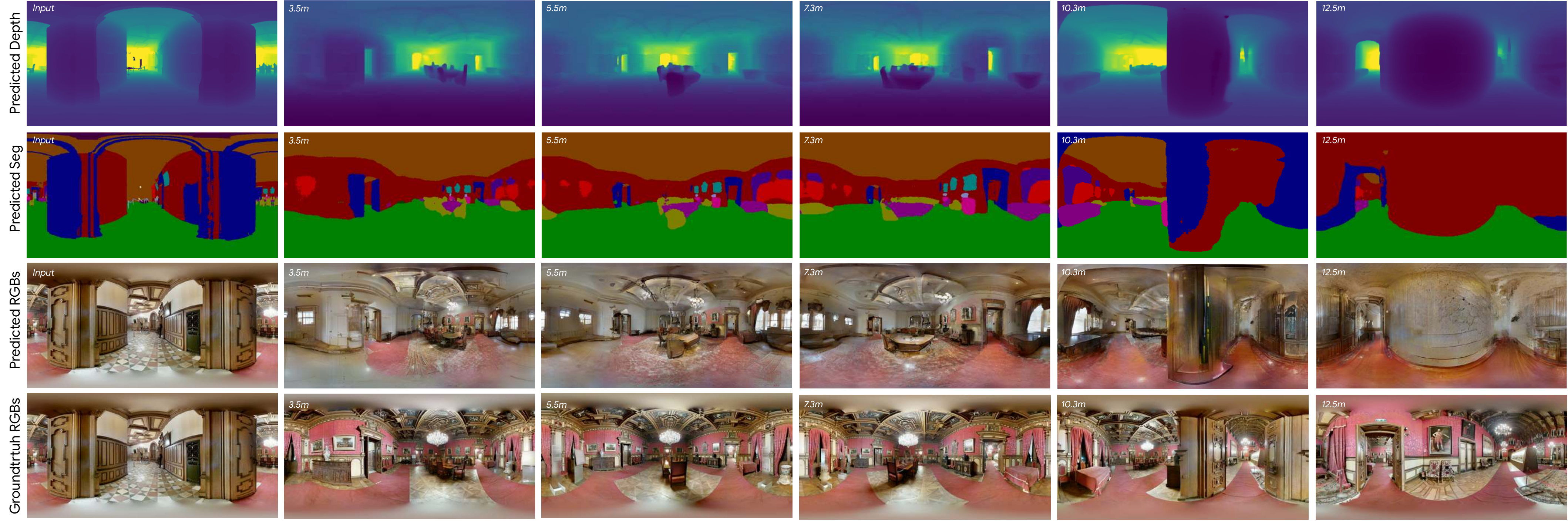}
    \vspace{-17pt}
    \caption{Despite the initial provided observation being quite distant, \pathdreamer is able to synthesize realistic segmentation and RGB details of the table, paintings, and doorway at 3.5m after moving closer. A plausible hallway scene is also synthesized at 10.3m and 12.5m.}
    \vspace{-5pt}
    \label{fig:qualitative_192}
    \end{subfigure}
\caption{Selected examples of predicted sequences from the \valunseen split using one ground truth observation as context.}
\label{fig:qualitative_valunseen}
\end{figure*}
\begin{figure*}[t]
\centering
    \begin{subfigure}{1.0\textwidth}
    \includegraphics[width=1.0\textwidth]{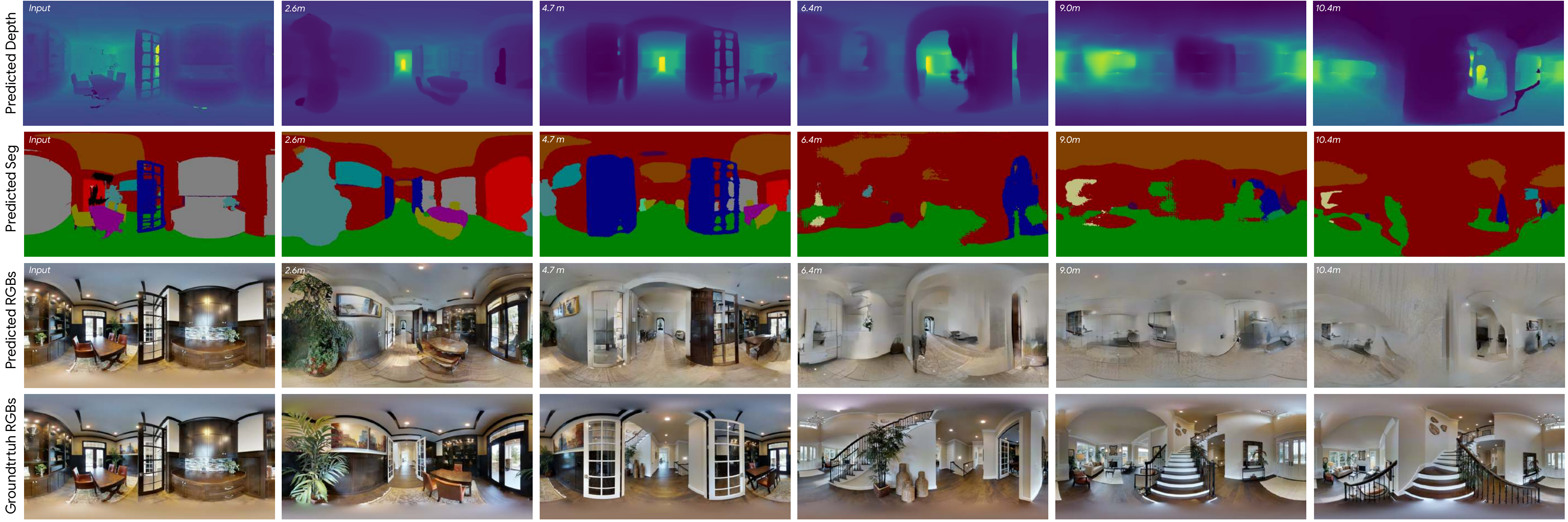}
    \vspace{-17pt}
    \caption{In this sequence, feature predictions in the segmentation space are generally preserved up to 4.7m. The predictions tend to diverge from the groundtruth after, due to the failure of the model to predict the existence of a stairwell at 4.7m}
    \vspace{3pt}
    \label{fig:random_99}
    \end{subfigure}

    \begin{subfigure}{0.97\textwidth}
    \includegraphics[width=1.0\textwidth]{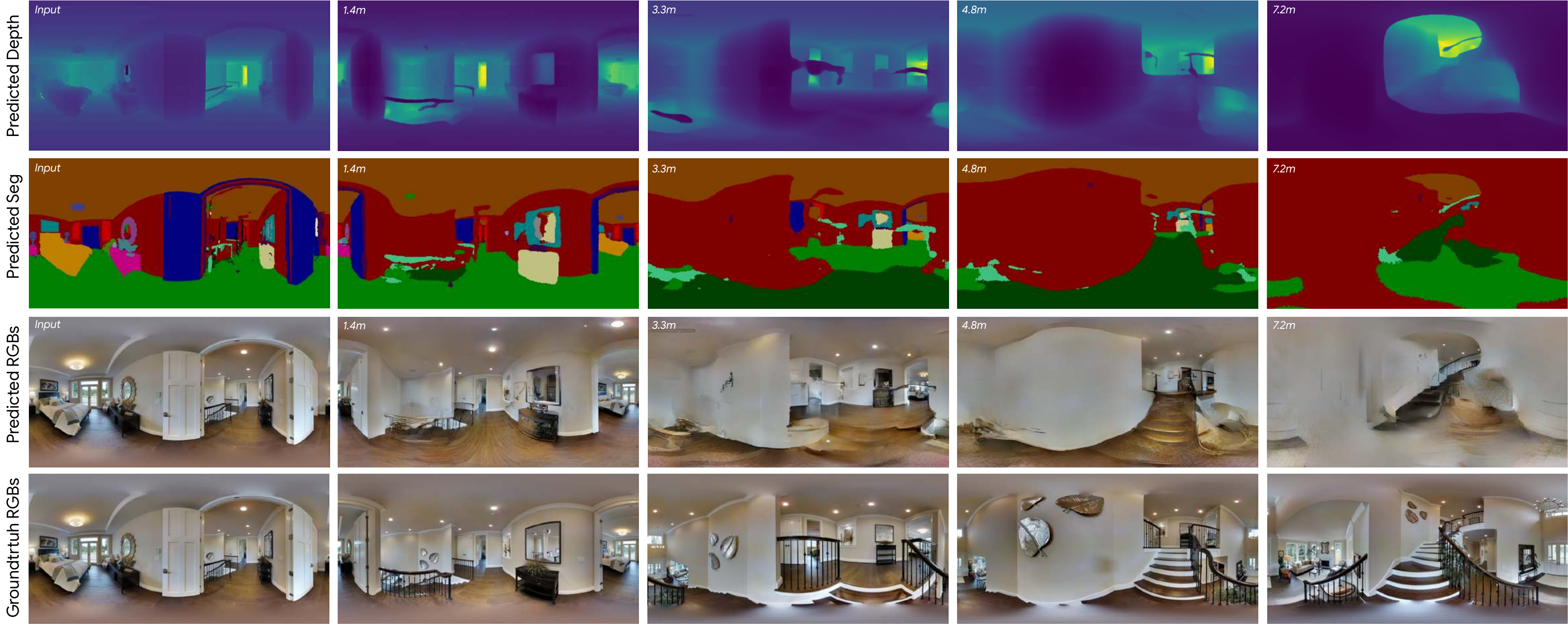}
    \vspace{-17pt}
    \caption{Model predictions in the segmentation space are generally accurate up to 4.8m. However, the corresponding RGB outputs are not as realistic, likely due to difficulties in generating intricate objects such as stair railings.}
    \vspace{3pt}
    \label{fig:random_272}
    \end{subfigure}

    \begin{subfigure}{1.0\textwidth}
    \includegraphics[width=1.0\textwidth]{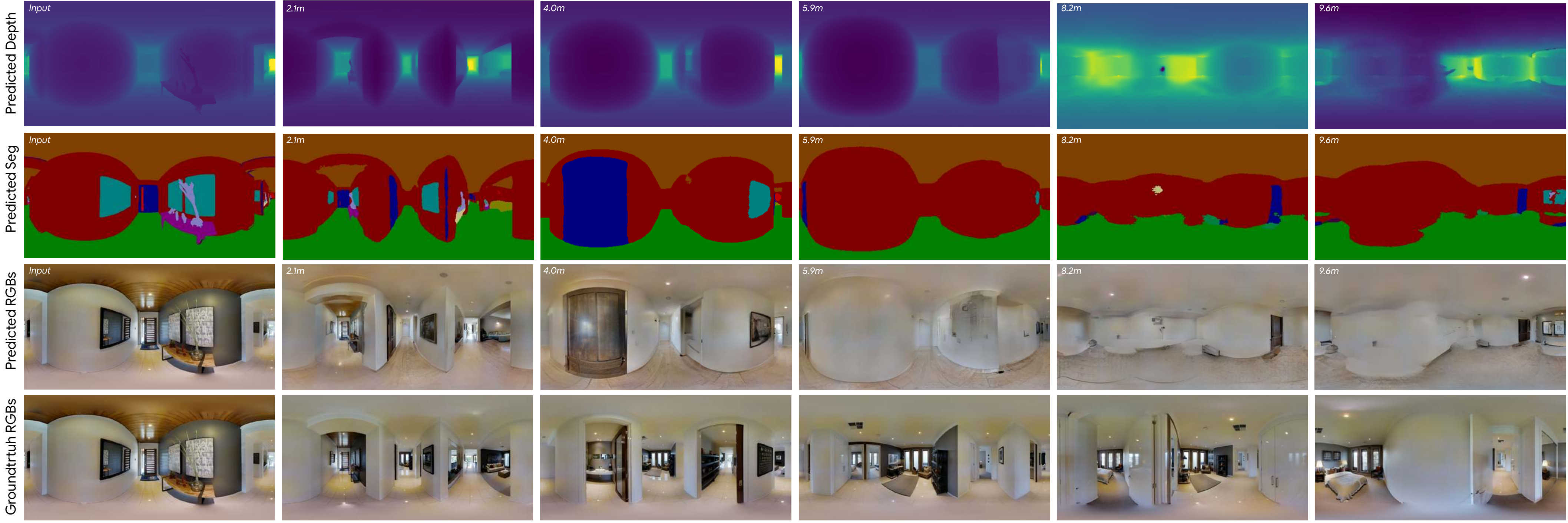}
    \vspace{-17pt}
    \caption{In this example, the model fails to predict the existence of the room entrance in the center of the panorama at 2.1m, which leads to the outputs eventually diverging from the ground truth. Despite this, predicted results remain generally plausible.}
    \label{fig:random_545}
    \end{subfigure}
\caption{Randomly selected prediction sequences from the \valunseen split using one ground truth observation as context.} \label{fig:random_unseen}
\end{figure*}

\begin{figure*}[t]
\centering

    \begin{subfigure}{1.0\textwidth}
    \includegraphics[width=1.0\textwidth]{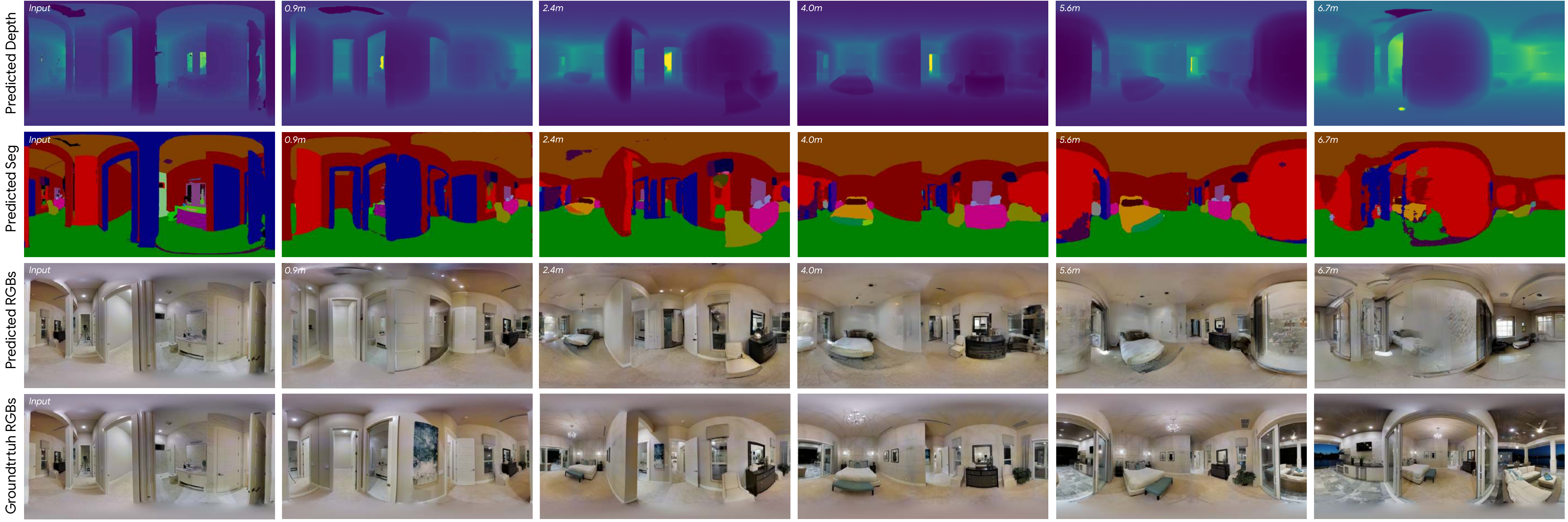}
    \vspace{-17pt}
    \caption{\pathdreamer is able to almost perfectly recreate the navigation sequence, despite only being provided with a single ground truth observation. Several details are missed in the RGB outputs, such as the sliding glass doors at 5.6m. This is likely due to the poor quality depth returns from glass doors in the training data.} \label{fig:qualitative_4}
    \vspace{3pt}
    \end{subfigure}

    \begin{subfigure}{1.0\textwidth}
    \includegraphics[width=1.0\textwidth]{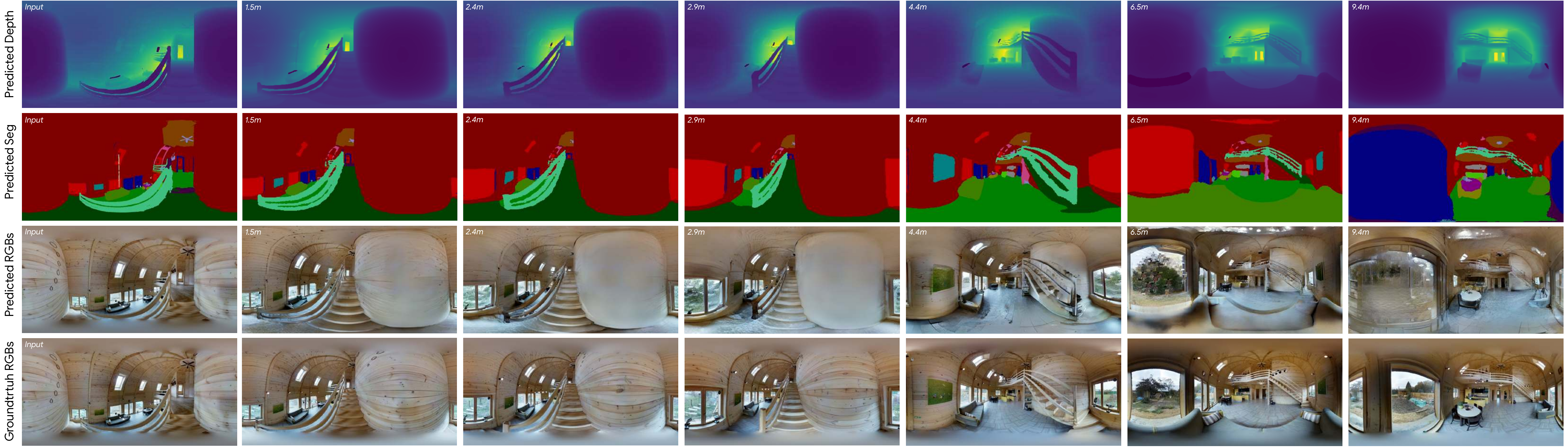}
    \vspace{-17pt}
    \caption{From the initial observation, \pathdreamer is able to recreate the scene from multiple perspectives: (1) the middle of the stairs at 1.5m, (2) the bottom of the stairs at 4.4m, and (3) from the corner of the room at 6.5m. The ability to accurately representing geometry is one of its strengths as a world model.} \label{fig:qualitative_10}
    \vspace{3pt}
    \end{subfigure}

    \begin{subfigure}{0.97\textwidth}
    \includegraphics[width=1.0\textwidth]{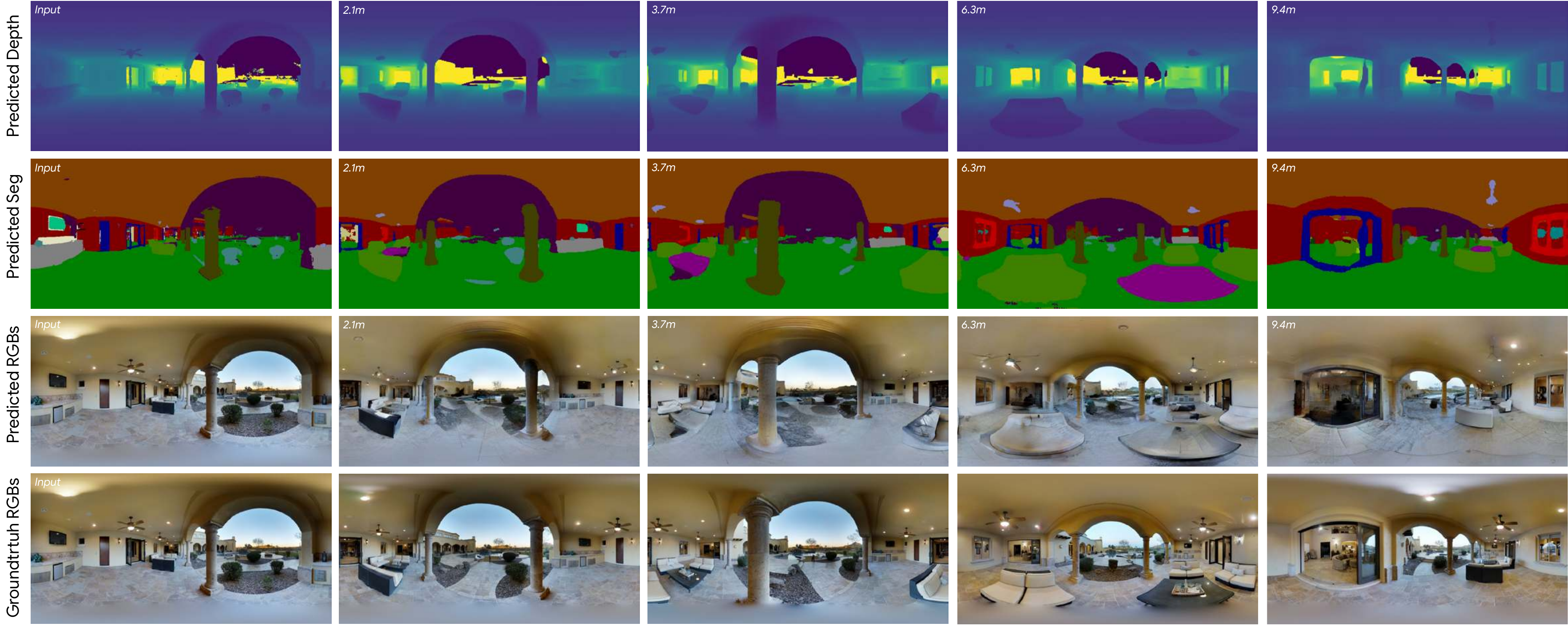}
    \vspace{-17pt}
    \caption{In this example, the outdoor scenery is accurately recreated due to the information from both the segmentation predictions and the projected RGB values. Some textural details are lost in the couch and table at 6.3m, which suggest the potential to improve results by training a stronger \rgbmodel.} \label{fig:qualitative_11}
    \end{subfigure}
\caption{Selected examples of predicted sequences from the \valseen split using one ground truth observation as context.}
\label{fig:qualitative_seen}
\end{figure*}
\begin{figure*}[t]
\centering
    \begin{subfigure}{0.92\textwidth}
    \includegraphics[width=1.0\textwidth]{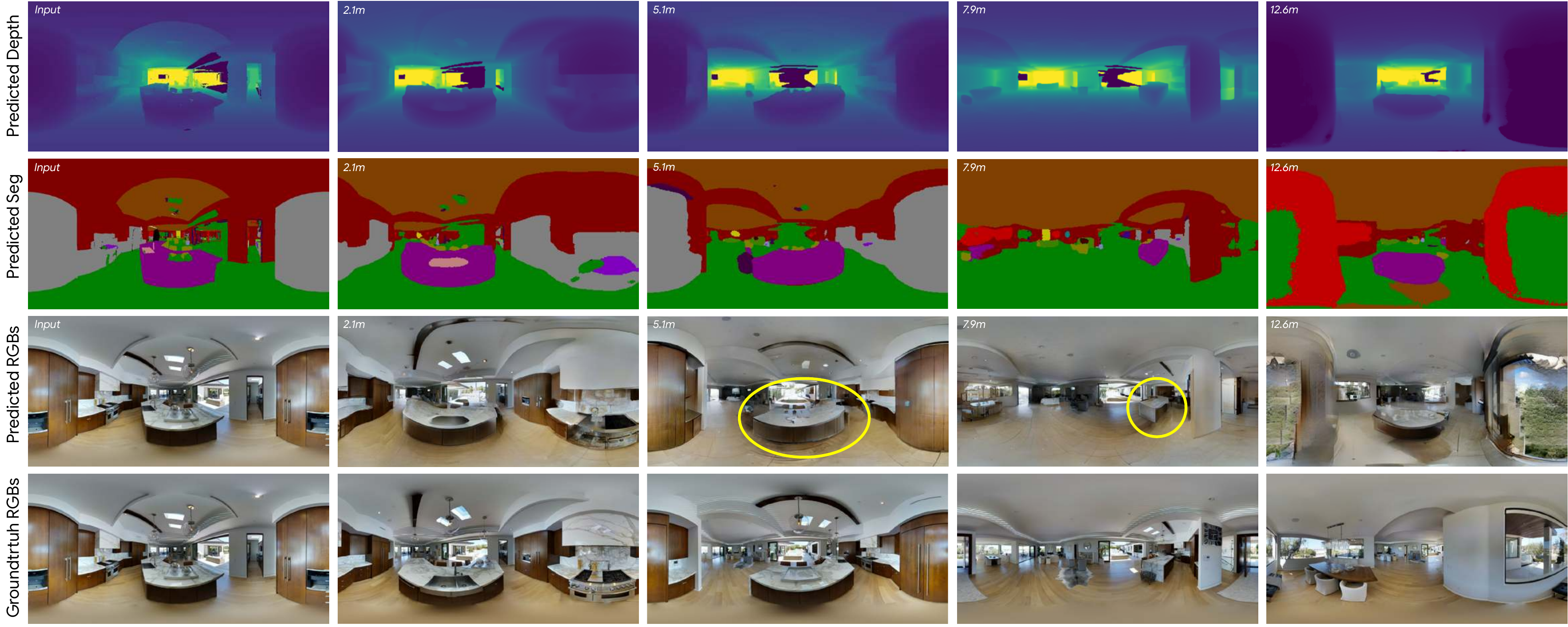}
    \vspace{-17pt}
    \caption{\pathdreamer is able to perform view synthesis for potentially unbounded viewpoint changes. Provided with an observation of the kitchen counter from one angle, realistic views can be synthesized from other perspectives (highlighted at 5.1m and 7.9m).} \label{fig:qualitative_random_seen_63}
    \vspace{0pt}
    \end{subfigure}

    \begin{subfigure}{0.92\textwidth}
    \includegraphics[width=1.0\textwidth]{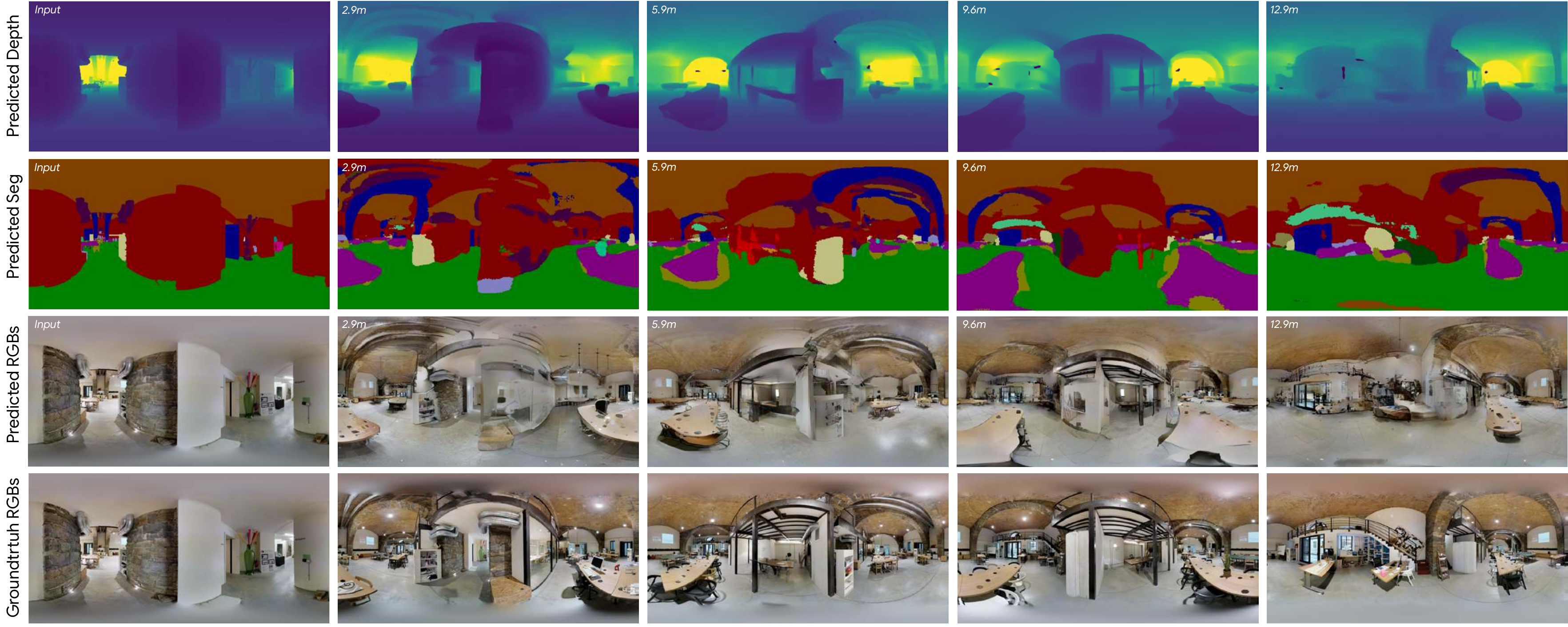}
    \vspace{-17pt}
    \caption{While \pathdreamer is able to maintain consistency of the generated semantic segmentations in this example, the complexity of the structures (e.g. arched ceiling beams) make RGB generation difficult, and the results are blurred in certain regions.} \label{fig:qualitative_random_seen_34}
    \vspace{0pt}
    \end{subfigure}

    \begin{subfigure}{0.92\textwidth}
    \includegraphics[width=1.0\textwidth]{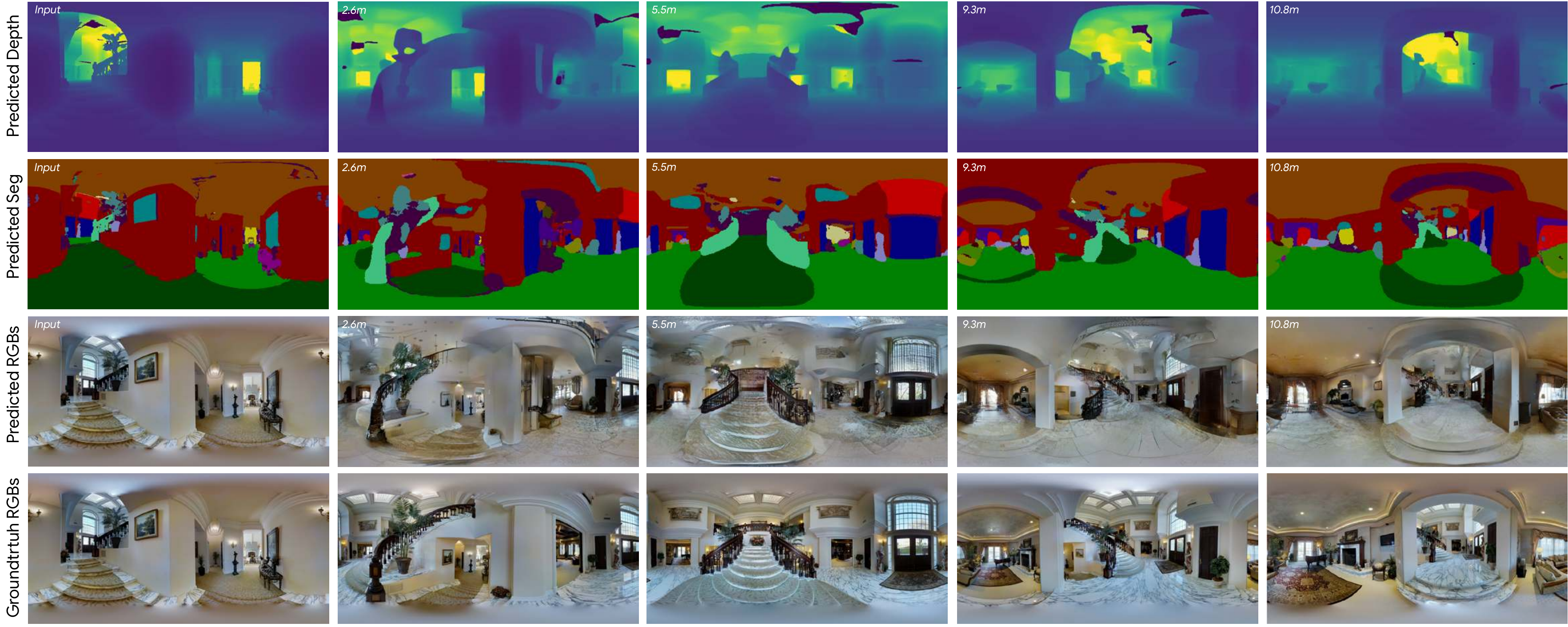}
    \vspace{-17pt}
    \caption{\pathdreamer is able to handle complex scene geometry, as shown in this example with multiple floors. While the overall scene looks fairly realistic, the \rgbmodel has some trouble with generating realistic looking stairs (e.g. at 5.5m).} \label{fig:qualitative_random_seen_242}
    \vspace{-5pt}
    \end{subfigure}
\caption{Random prediction sequences from the \valseen split using one ground truth observation as context.}
\label{fig:random_seen}
\end{figure*}

\subsection{\valunseen Results}
We present additional qualitative results generated by the \pathdreamer model, on the \valunseen split of the R2R dataset. These environments are not seen by the model during training, and act as a test of the generalization ability of \pathdreamer.
Figure~\ref{fig:qualitative_valunseen} presents cherry-picked examples of \pathdreamer generated sequences and Figure~\ref{fig:random_unseen} presents randomly chosen examples.

\subsection{\valseen Results}
The \valseen split contains novel navigation trajectories for environments seen in the training split. As described in the main paper, the \segmodel and \rgbmodel models perform significantly better on \valseen, since they can memorize the training environments with less generalization required. We observe that generation results maintain high fidelity even at large distances from the location of the input observation. 
Figure~\ref{fig:qualitative_seen} presents cherry-picked examples and Figure~\ref{fig:random_seen} presents random examples.

\subsection{Noise Interpolation}
\begin{figure*}[t]
\centering
    \begin{subfigure}{1.0\textwidth}
    \includegraphics[width=1.0\textwidth]{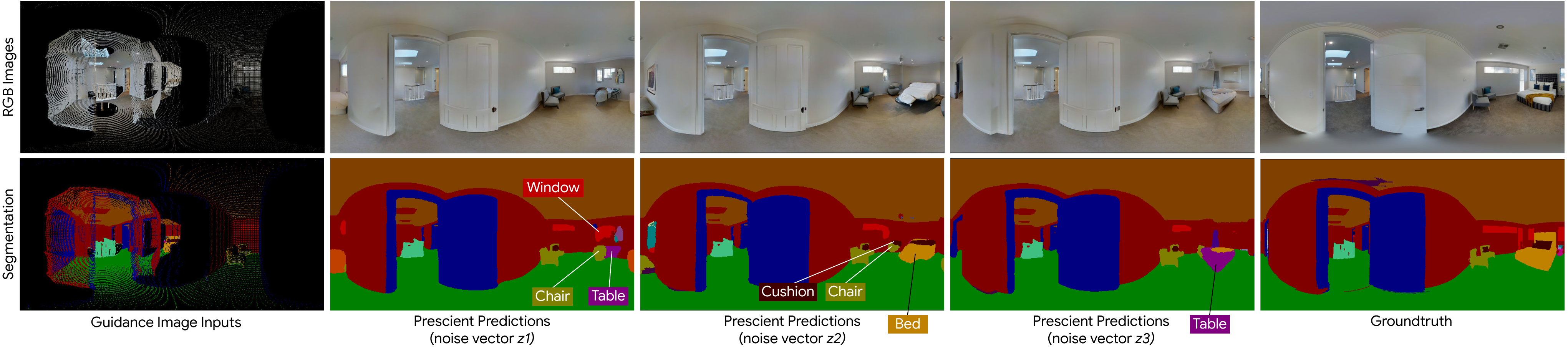}
    \vspace{-15pt}
    \caption{Several plausible scene layouts are predicted for the unknown region of the image on the right (indicated in black in the guidance inputs). Notably, \pathdreamer is able to synthesize a complete set of a bed, cushion, and chair in the second example (noise vector $z_2$).} \label{fig:qualitative_noise_15}
    \vspace{5pt}
    \end{subfigure}

    \begin{subfigure}{1.0\textwidth}
    \includegraphics[width=1.0\textwidth]{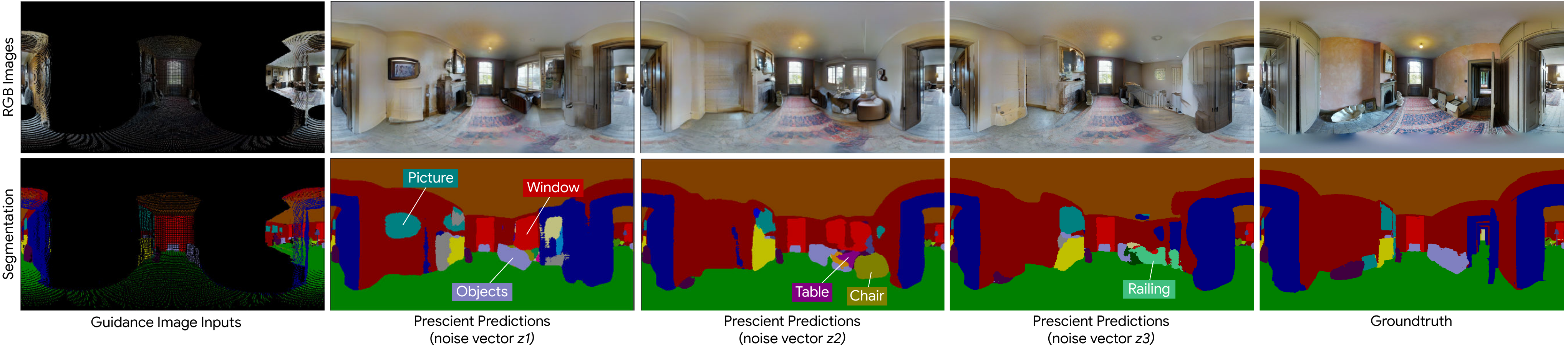}
    \vspace{-15pt}
    \caption{\pathdreamer predicts several plausible scenes: (1) a room with a painting on the left wall, (2) a room without a painting, but with a table and chair on the right, and (3) a room without a table or chair, but with stairs leading down.} \label{fig:qualitative_noise_72}
    \vspace{5pt}
    \end{subfigure}

    \begin{subfigure}{1.0\textwidth}
    \includegraphics[width=1.0\textwidth]{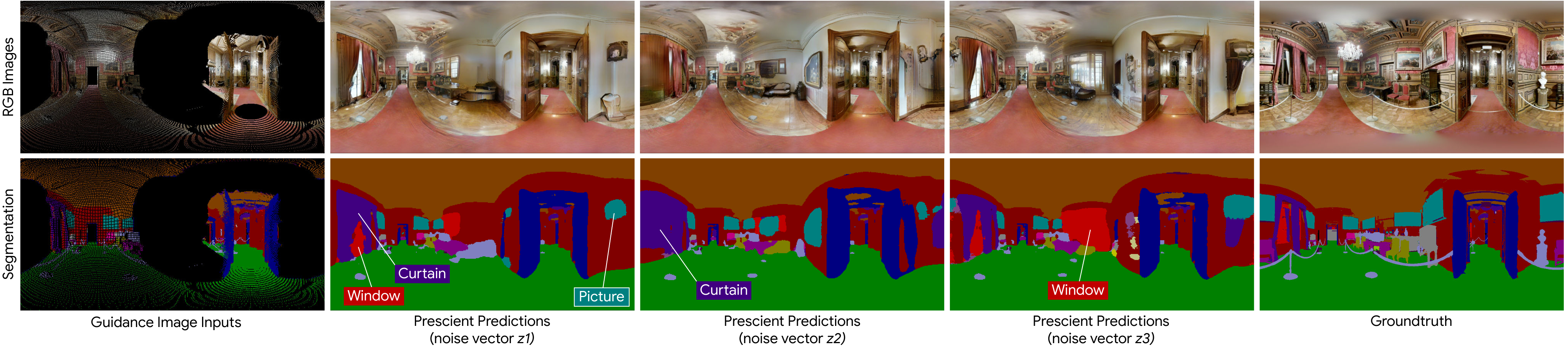}
    \vspace{-15pt}
    \caption{Several varying but plausible scenes are considered by \pathdreamer: (1) a window with curtains open at the side, (2) a window covered by curtains, and (3) a large floor to ceiling window on the right of the room.} \label{fig:qualitative_noise_104}
    \end{subfigure}
\caption{Diverse and semantically plausible predictions can be sampled for scenes containing previously unseen regions. Shown here are several selected examples, each consisting of three alternative room reveals and the corresponding ground truth (from \valunseen).}
 \label{fig:selected_noise}
\end{figure*}

\begin{figure*}[t]
\centering
    \begin{subfigure}{1.0\textwidth}
    \includegraphics[width=1.0\textwidth]{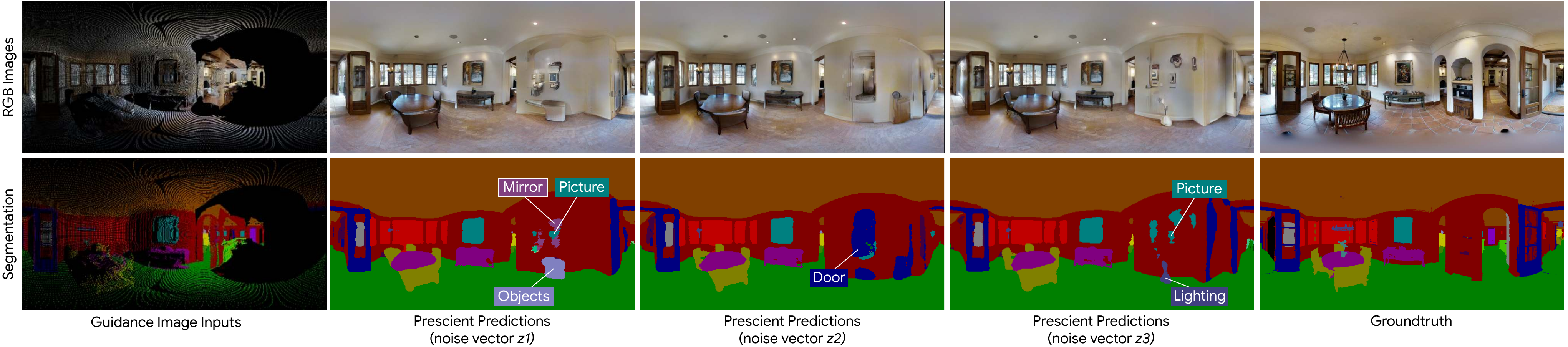}
    \vspace{-15pt}
    \caption{A wall is correctly predicted to exist on the right of the image. Varying layouts and decorations for the wall are proposed, which all create valid segmentation and RGB generation results.} \label{fig:qualitative_noise_234}
    \vspace{5pt}
    \end{subfigure}

    \begin{subfigure}{1.0\textwidth}
    \includegraphics[width=1.0\textwidth]{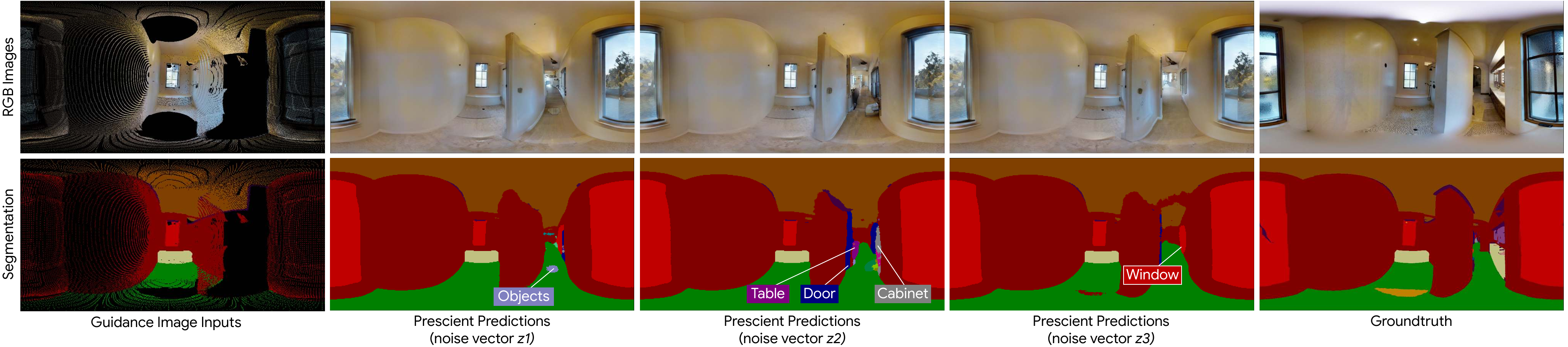}
    \vspace{-15pt}
    \caption{In this example, the missing region of the image is correctly predicted to be a hallway. Several plausible variations of the hallway are synthesized, each containing objects that are likely to be present.} \label{fig:qualitative_noise_463}
    \vspace{5pt}
    \end{subfigure}

    \begin{subfigure}{1.0\textwidth}
    \includegraphics[width=1.0\textwidth]{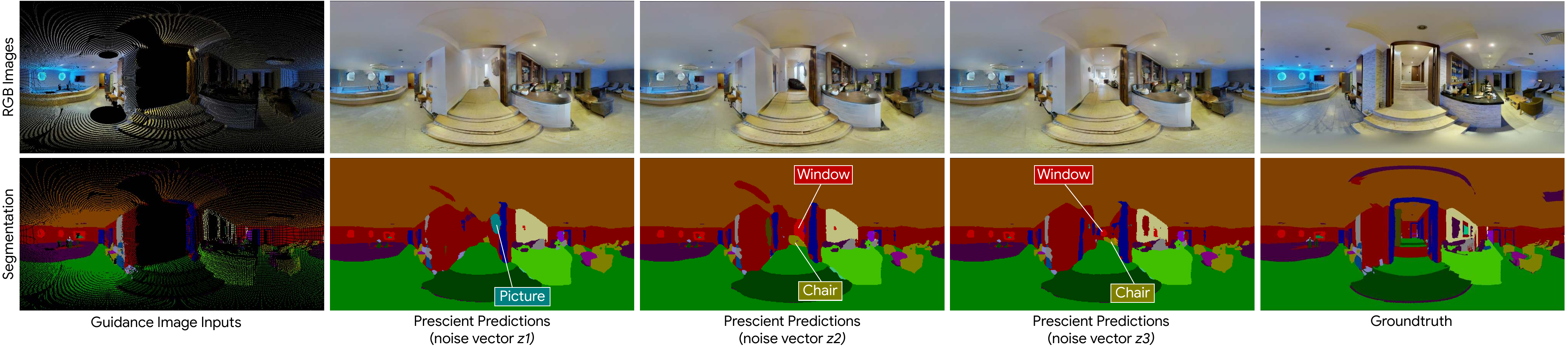}
    \vspace{-15pt}
    \caption{This example contains stairs leading to a hallway, which is correctly predicted by the model. Several potential layouts are synthesized, although none of them fully synthesize the entire door frame present in the ground truth example.} \label{fig:qualitative_noise_478}
    \end{subfigure}
\caption{Random examples of generation results when sampled with different noise vectors.}
 \label{fig:random_noise}
\end{figure*}

As described in the main paper, the \segmodel is able to generate alternative, diverse, \textit{room reveals} for a given scene. In areas without guidance pixels (i.e., in areas of the environment not seen from a previous view), interpolating over different noise vectors produces diverse and plausible outcomes.
Figure~\ref{fig:selected_noise} presents cherry-picked examples of \pathdreamer sequences when conditioned on different noise vectors. Figure~\ref{fig:random_noise} contains additional examples of randomly selected sequences conditioned on random noise vectors.

\section{Generated Videos}
In addition to image generation, \pathdreamer is capable of generating continuous video sequences, simply by sequencing generated images with small viewpoint changes. We provide videos displaying \pathdreamer generated results for several unseen environments. We also compare \pathdreamer's rendering quality using multiple observations to the output of the Habitat simulator using mesh-based rendering, illustrating a favorable comparison in quality. We refer readers to the YouTube link\footnote{\url{https://youtu.be/StklIENGqs0}} for the video results.

\section{Implementation Details}
\paragraph{\segmodel} We trained this model with a batch size of 64, over 50 epochs. For the first 30 epochs the model is trained with teacher forcing, i.e., the previous observation that is used as input at each step is the ground-truth previous observation. During the teacher forcing stage, the number of ground truth context frames used for a path of length $L$ is decayed uniformly from $L-1$ to $1$. After 30 epochs, we switch to the recurrent setting in which the model's previous prediction is used as the previous observation at each step during the rollout (similar to our setup at inference time). We use the Adam optimizer with parameters $\beta_1 = 0.9$ and $\beta_2 = 0.999$. We apply a learning rate which starts at $1e^{-4}$ and warms up to $2e^{-4}$ uniformly over 10 epochs.

\paragraph{\rgbmodel} This model was trained with a batch size of 128 over 500 epochs. We use the Adam optimizer with parameters $\beta_1 = 0.5$ and $\beta_2 = 0.999$, and a learning rate of $2e^{-4}$ for both the generator and the discriminator. During training, the discriminator is trained for 2 steps for each generator training step. Following standard practice, at inference time we applied an exponential moving average (EMA) to the generator weights with 0.999 decay.  For the choice of loss weights, we set $\lambda_{\text{GAN}} = 1$, $\lambda_{\text{VGG}} = 0.07$, and $\lambda_{\text{FM}} = 0$. We found experimentally that excluding the feature matching loss speeds up training throughput and did not have a significant effect on the results.

\paragraph{Evaluation Details}
We compute the FID score\footnote{We used \url{https://github.com/mseitzer/pytorch-fid} for computing results.} using 10,000 random samples for each prediction sequence step. As the R2R validation sets contain 783 and 340 sequences for \valunseen and \valseen respectively, we perform data augmentation with random horizontal roll and flips to acquire 10,000 samples.

We run evaluation every 2000 training steps, and select the best checkpoint on the \valseen and \valunseen set for reporting results for the respective split. We note that training time is generally significantly longer on \valseen as compared to \valunseen. Due to the benefit of overfitting on the training set for \valseen, performance continues to improve on even as generalization performance on \valunseen deteriorates.


\end{document}